\documentclass{article} 
\usepackage{iclr2021_conference}
\preprint
\usepackage{times}
\usepackage[utf8]{inputenc} 
\usepackage[T1]{fontenc}    
\usepackage{url}            
\usepackage{booktabs}       
\usepackage{nicefrac}       
\usepackage{microtype}      

\usepackage{amsmath}
\usepackage{amsfonts}       
\usepackage{amssymb}
\usepackage{amsthm}
\usepackage{nicefrac}
\usepackage[mathscr]{euscript}

\usepackage{algorithm}
\usepackage{algpseudocode}

\usepackage{xcolor}
\definecolor{fg}{cmyk}{0.0, 0.0, 0.02, 0.03}
\definecolor{cyberblue}{HTML}{0016EE}
\definecolor{twilightblue}{HTML}{363b74}
\definecolor{twilightpink}{HTML}{ef4f91}
\definecolor{twilightbluepurple}{HTML}{4d1b7b}
\definecolor{twilightpurple}{HTML}{673888}

\usepackage{hyperref}
\usepackage{cleveref}
\hypersetup{
  colorlinks=true,
  citecolor=twilightblue,
  linkcolor=twilightpink,
  linktocpage=true,
  urlcolor=twilightpurple,
}

\usepackage{todonotes}
\setuptodonotes{
  linecolor=twilightbluepurple,
  backgroundcolor=twilightbluepurple,
  textcolor=fg,
  bordercolor=white,
}

\usepackage{graphicx}
\usepackage{siunitx}
\usepackage{subcaption}
\usepackage[export]{adjustbox}

\usepackage{natbib}
\setcitestyle{authoryear,open={(},close={)}}

\newcommand{\M}{\mathcal{M}} 
\newcommand{\A}{\mathcal{A}} 
\newcommand{\X}{\mathcal{X}} 
\newcommand{\Op}{\mathcal{O}} 
\newcommand{\ah}{\hat{A}} 

\newcommand{\defequal}{\mathrel{\overset{\makebox[0pt]{\mbox{\tiny def}}}{=}}}
\newcommand{\ubcomment}[2]{\underbrace{#1}_{\mbox{\small #2}}}
\newcommand{\sample}[1]{\ubcomment{#1}{sample}}

\newcommand{\E}{\mathbb{E}} 
\newcommand{\1}{\mathbb{I} } 
\newcommand{\indic}[1]{\1_{#1}} 

\newcommand{\I}{\mathscr{I}} 
\newcommand{\Io}{\I^o}
\newcommand{\po}{P^o}

\newcommand{\bo}{{\beta^o}}
\newcommand{\deriv}{\nabla_{\theta_{\beta}}}

\newcommand{\pio}{{\pi^o}}

\newcommand{\qo}{Q_{\Op}}
\newcommand{\vo}{V_{\Op}}
\newcommand{\uo}{U_{\Op}}

\newcommand{\pop}{p_{\Op}}
\newcommand{\ph}{\hat{\pop}}

\newcommand{\etah}{\hat{\eta}}

\newcommand{\cent}{c_{H}}
\newcommand{\centb}{c_{H_\beta}}
\newcommand{\cvic}{c_{\textrm{VIC}}}

\newcommand{\rvic}{R_\textrm{VIC}}
\newcommand{\kvic}{K_\textrm{VIC}}

\newtheorem{theorem}{Theorem}
\newtheorem{lemma}{Lemma}
\newtheorem{assumption}{Assumption}
\newtheorem{proposition}{Proposition}

\newcounter{num}

\newcommand{\cutsectionup}{\vspace*{-0.06in}}
\newcommand{\cutsectiondown}{\vspace*{-0.05in}}
\newcommand{\cutsubsectionup}{\vspace*{-0.04in}}
\newcommand{\cutsubsectiondown}{\vspace*{-0.03in}}
\newcommand{\cutparagraphup}{\vspace*{-0.04in}}

\title{Learning Diverse Options via \\ InfoMax Termination Critic}
\author{
  Yuji Kanagawa
  \thanks{Work is partly done at the University of Tokyo} \\
  Okinawa Institute of Science and Technology \\
  \texttt{yuji.kanagawa@oist.jp} \\
  \And
  Tomoyuki Kaneko \\
  The University of Tokyo \\
  \texttt{kaneko@graco.c.u-tokyo.ac.jp}
}

\begin{document}
\maketitle
\begin{abstract}
We consider the problem of autonomously learning reusable temporally extended actions, or options, in reinforcement learning.
While options can speed up transfer learning by serving as reusable building blocks, learning reusable options for unknown task distribution remains challenging.
Motivated by the recent success of mutual information (MI) based skill learning, we hypothesize that more diverse options are more reusable.
To this end, we propose a method for learning termination conditions of options by maximizing MI between options and corresponding state transitions.
We derive a scalable approximation of this MI maximization via gradient ascent, yielding the InfoMax Termination Critic (IMTC) algorithm.
Our experiments demonstrate that IMTC significantly improves the diversity of learned options without extrinsic rewards, combined with an intrinsic option learning method.
Moreover, we test the reusability of learned options by transferring options into various tasks, confirming that IMTC helps quick adaptation, especially in complex domains where an agent needs to manipulate objects.
\end{abstract}






\cutsectionup
\section{Introduction} \label{section:intro}
\cutsectiondown
Behavior learning from environmental interaction is a fundamental problem in artificial intelligence and robotics.
In recent years, combined with deep neural networks (DNN), reinforcement learning (RL) has successfully learned complex behaviors guided by human-designed reward functions \citep{Heess2017EmLoco, OpenAI2019Rubik}.
To reuse trained agents across multiple reward functions, or tasks, abstracting a course of action as a higher-level building block \citep{Barto2003HRL,Barto2013BH} would be useful.
The options framework~\citep{Sutton1999Option} is a representative formulation of action abstraction in RL, where each option consists of an intra-option policy and its termination function.
Options have the potential of accelerating learning in new tasks \citep{Brunskill2014PAC}, serving as reusable behavioral blocks.
For example, if a wheeled robot already had learned to lean and turn left and right, it can learn to drive on unfamiliar roads faster than learning from scratch.
However, discovering reusable options remains challenging due to the difficulty of define and measure reusability a priori.
Although we can measure the reusability of options when we know the task distribution is known (e.g., in a Bayesian way \citep{Solway2014Optimal}), it is difficult to measure it without prior knowledge.
Hence, in such cases, we need a reasonable heuristic assumption about future tasks.

In this paper, we hypothesize that diversifying options improves the reusability of options, assuming that tasks are uniformly distributed over the state space.
We argue that this is a reasonable heuristic because diverse options are expected to work well for any task to some degree.
The next problem is in measuring the diversity of options.
The most intuitive strategy would be visiting all states as equally as possible.
However, in environments with continuous or large state spaces, targetting all states could be computationally impossible or does not make much sense.
As an alternative, in the context of autonomous skill acquisition that considers learning a set of policies without termination conditions, mutual information (MI) maximization has been widely applied for learning diverse policies \citep{Eysenbach2019DIAYN, Baumli2021RVIC, Choi2021VE}.
MI maximization is advantageous in that it has tractable approximations with probabilistic models learned by DNN, enabling it to scale up to more complex domains.
\begin{figure}[t]
  \centering
  \begin{subfigure}[t]{2.5cm}
    \includegraphics[width=2.5cm]{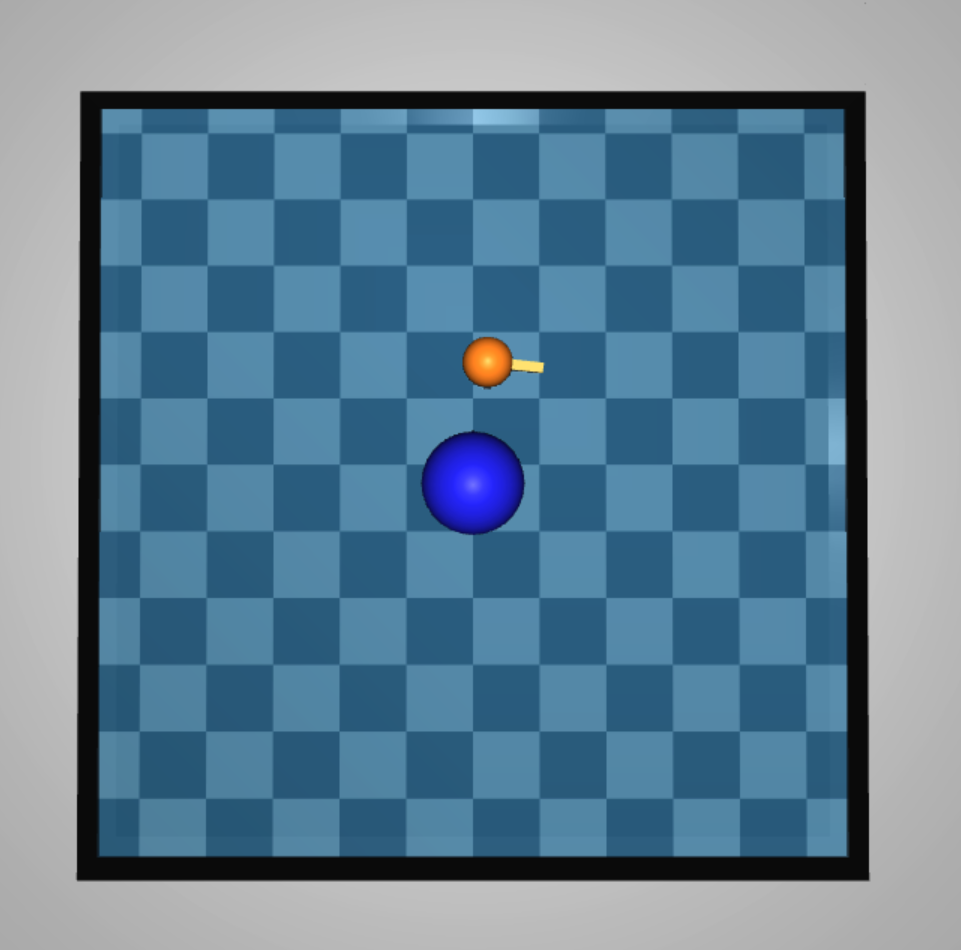}
  \end{subfigure}
  \quad
  \begin{subfigure}[t]{10cm}
    \includegraphics[width=10cm]{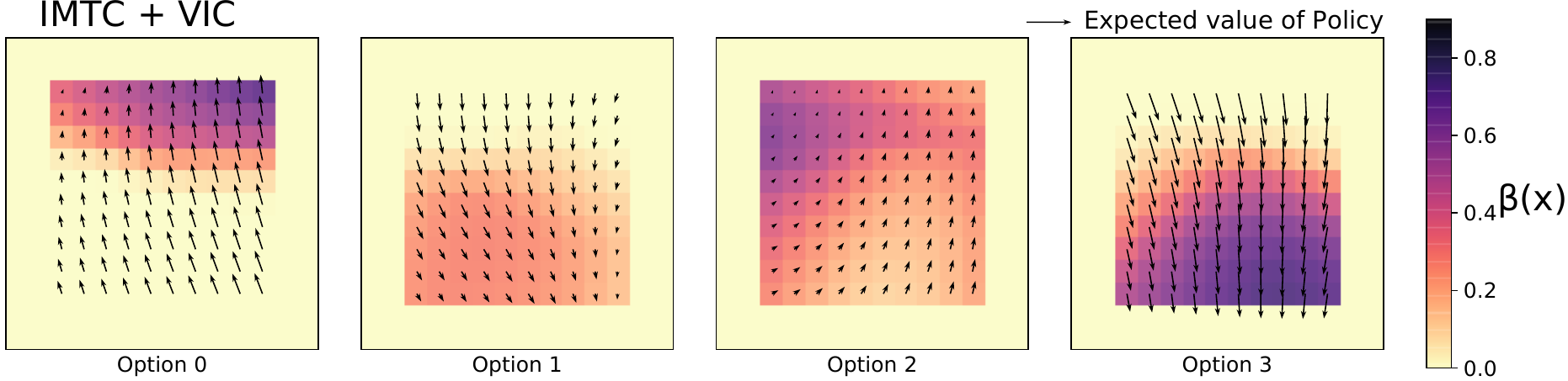}
    \includegraphics[width=10cm]{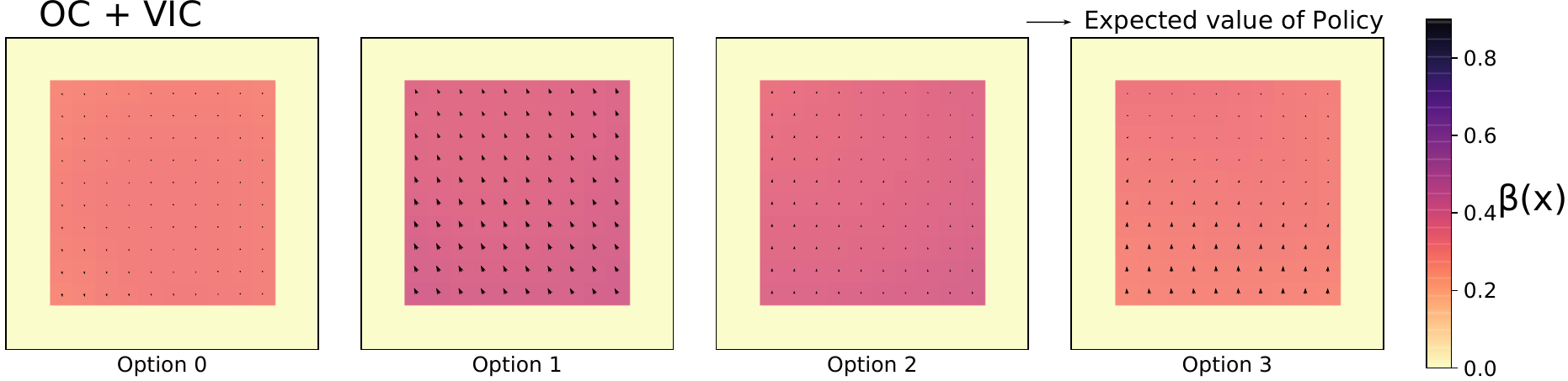}
  \end{subfigure}
  \vspace{-2mm}
  \caption{
    \textbf{Left:} \texttt{PointBilliard} environment.
    \textbf{Right:} Learned options by IMTC (upper row) and OC (lower row).
    Arrows show policies, and heatmaps show termination conditions of options.
  }
  \vskip -0.1in
  \label{figure:point-billiard-options}
\end{figure}

To this end, we propose maximizing MI between options and their terminating states conditioned by a starting state for learning diverse options.
Maximizing this MI diversifies the destinations of an agent when it chooses different options at a certain state while keeping the transition as deterministic as possible, leading to diverse and meaningful options.
The same objective is employed by \citet{Gregor2017VIC} for learning sub-policies, but we aim to train termination conditions of options as well.
We derive an unbiased estimator of the gradient of this MI with respect to termination conditions using the termination gradient theorem \citep{Harutyunyan2019TC}.
As a key contribution, we reformulate the estimated gradient using Bayes' rule and derive a tractable approximation of this MI maximization with a leaned classification model, yielding the InfoMax Termination Critic (IMTC) algorithm.
We implement IMTC on the Option Critic Architecture (OC, \citet{Bacon2017OC}) and Proximal Policy Optimization (PPO, \citet{Schulman2017PPO}) with a newly proposed advantage estimation for option learning.
In experiments, we qualitatively demonstrate that IMTC successfully learns diverse and meaningful options in reward-free RL \citep{Jin2020RewardFree} setting, combined with VIC \citep{Gregor2017VIC}.
\Cref{figure:point-billiard-options} shows the learned options by IMTC and OC in \texttt{PointBilliard} domain.
We can see that options learned by IMTC are clearly separated and different.
Moreover, we test the reusability of learned options in task adaptation experiments: we first pre-train options without rewards and then transfer them on various tasks. We show that IMTC helps quick adaptation to specific tasks, especially in complex domains where an agent needs to manipulate objects.
\cutsectionup
\section{Background and Notation} \label{section:background}
\cutsectiondown
We assume the standard RL setting in the Markov decision process~(MDP), following~\citet{Sutton2018RL}.
MDP $\M{}$ consists of a tuple $(\X{}, \A{}, p, r)$, where $\X{}$ is the finite set of states, $\A{}$ is the finite set of actions, $p: \X\times\A\times X \rightarrow [0, 1]$ is the state transition
function, $r: \X\times\A \rightarrow [r_\textrm{min}, r_\textrm{max}]$ is the reward function, where $r_\textrm{min}, r_\textrm{max}$ are minimum and maximum.
We use $X_t$ and $A_t$ for denoting random variables of the state and action experienced at time $t$.
We define $G \defequal{} \sum_{t=0}^{\infty}\gamma^t R_{t}$ as discounted cumulative return, where $R_t = r(X_t, A_t)$ is the reward received at time $t$ and $0<\gamma<1$ is the discount factor.
We consider maximizing expected discounted return $\E \left[G|\pi\right]$.
As an agent, we consider a memoryless policy $\pi: \X \times \A \rightarrow [0, 1]$.
$\pi$ induces two value functions: action-value function $ Q^\pi (x, a) \defequal{} \E_\pi \left[G|X_0 = x, A_0 = a, \pi \right]$ and state-value function $ V^\pi(x) \defequal{} \sum_{a}\pi(a|x) Q^\pi(x, a)$.
We use $d^\pi$ for denoting an agent's average occupancy measure $d^\pi (x) = {\displaystyle \scriptsize \lim_{n \to \infty}} \frac{\E\left[ \sum_{t=0}^n \indic{X_t=x} \right]}{n}$, where $\indic{}$ is the indicator function.

Assuming that $\pi$ is differentiable by the policy parameters $\theta_\pi$, the policy gradient (PG) method~\citep{Williams1992REINFORCE} maximizes $G^\pi$ by updating $\theta_\pi$ via gradient ascent.
A common formulation of PG estimates the gradient by
$ \nabla_{\theta_\pi} G^\pi = \E_{x, a, \pi} \bigg[
    \nabla_{\theta_\pi} \log \pi(a|x) \ah (x, a)
  \bigg] $,
where $\ah (x, a)$ is an estimation of the advantage function
$A^\pi (x, a) \defequal{} Q^\pi(x, a) - V^\pi(x)$.
Among many variantes of PG methods, we implemented our method on PPO~\citep{Schulman2017PPO}.
At each gradient step, PPO updates $\theta_\pi$ to maximize $\textrm{clip}(\frac{\pi(a|x)}{\pi_\textrm{old}(a|x)}\ah,-\epsilon, \epsilon)$, where $\textrm{clip}(x,-\epsilon,\epsilon)=\max(-\epsilon,\min(\epsilon,x))$ and $\pi_\textrm{old}$ is the policy before being updated.
This clipping heuristics prevents $\pi$ from updating too rapidly.

\cutparagraphup
\paragraph{Options Framework}
Options~\citep{Sutton1999Option} provide a framework for formulating temporally abstracted actions in RL.
An option $o \in \Op$ consists of a tuple $(\Io, \bo, \pio)$, where $\Io \subseteq{} \X$ is the initiation $\bo : \X \rightarrow{} [0, 1]$ is a termination function, and $\pio$ is an \textit{intra-option} policy.
Following related studies~\citep{Bacon2017OC,Harutyunyan2019TC}, we assume that $\Io = \X$ focuses on learning $\bo$ and $\pio$.
We let $\mu: \X \times \Op \rightarrow [0, 1]$ denote a policy over options.
A typical RL agent sample $A_t$ from $\pi(\cdot|X_t)$.
Analogously, at time $t$, an RL agent with options first samples a termination $T_t$ from $\beta^{O_t}(X_t)$.
If $T_t = 1$, $O_{t + 1}$ is sampled from $\mu(\cdot|X_t)$ and if not, the current option remains the same.
The next action $A_t$ is sampled from $\pi^{O_{t + 1}}(\cdot|X_t)$.
For option learning methods, we use $\pi$ to denote the resulting policy induced by $\mu$, $\cup_{o \in \Op} \pio$, and $\cup_{o \in \Op} \bo$.

\cutparagraphup
\paragraph{Option value functions}
With options, we have three option-value functions $\qo$, $\vo$, and $\uo$.
$\qo$ is the option-value function denoting the value of selecting an option $o$ at state $x$ defined by
$\qo(x, o) \defequal{} \E \left[G | X_0 = x, O_0 = o \right]$.
Similar to the relationship between $Q^\pi$ and $V^\pi$, we let $\vo$ denote the marginalized option-value function $\vo(x) \defequal{} \sum_{o} \mu(o|x) \qo(x, o)$.
$\uo(x, o) \defequal{} (1 - \bo(x))\qo(x, o) + \bo(x) \vo(x)$ is called the option-value function \textit{upon arrival}~\citep{Sutton1999Option} and denotes the value of reaching a state $x$ with $o$ and not having selected the new option.
We use these notations in \Cref{subsection:implementation:adv}.

\cutparagraphup
\paragraph{Termination gradient theorem}
Analogously to $p$, we let $\po: \X\times\Op\times X \rightarrow [0, 1]$ denote the state transition probability induced by options.
When an agent is at $x_s$ and having an option $o$, the probability that $o$ ends at $x_f$ is given by:
\begin{align}
\po (x_f|x_s) &= \bo(x_f)\indic{x_f=x_s} + (1-\bo(x_s))
 \sum_{x}p^{\pio}(x|x_s)\po(x_f|x),
\end{align}
where $p^\pio$ is the \textit{policy-induced} transition function $p^\pio(x'|x) \defequal{} \sum_{a\in\A}\pio(a|x)P(x'|x,a)$.
Here we assume that all options eventually terminate, such that $\po$ is a valid probability distribution over $x_f$.
Interestingly, $\po$ is differentiable with respect to the parameter of $\bo$.
\citet{Harutyunyan2019TC} introduced the termination gradient theorem:
\begin{theorem} \label{theorem:termination-gradient}
Let $\bo$ be parameterized by $\theta_\beta$, and let $\ell_\bo$ denote the logit of $\bo$, i.e., $\ell_\bo = \log(\frac{\bo(x)}{1 - \bo(x)})$.
We have
\begin{align} \label{eq:termination-gradient}
 \deriv \po(x_f|x_s) = \sum_{x} \po(x|x_s)
 \deriv \ell_{\bo}(x)(\indic{x_f=x} - \po(x_f|x)).
\end{align}
\end{theorem} \vspace*{-0.02in}
We use this theorem to derive an unbiased estimator of our target gradient.
\cutsectionup
\section{InfoMax Termination Critic} \label{section:methods}
\cutsectiondown

We now present the InfoMax Termination Critic~(IMTC) algorithm.
To learn diverse options, we propose to maximizes the following MI at each state $x_s$:
\begin{align} \label{eq:infomax}
  I(X_f; O| x_s) = H(X_f|x_s) - H(X_f|x_s, O) = H(O|x_s) - H(O|X_f, x_s),
\end{align}
where $I$ denotes the MI $I(A; B |c) = H(A|c) - H(A|B, c)$, $H$ denotes entropy, and $O$ is a random variable denoting options experienced by an agent.
Precisely, we let $\eta$ denote the probability of having an option $o$ when leaving a state $x$: $\eta(o|x) = \bo(x) \mu(o|x) + (1 - \bo(x)) \sum_{x' \in \X} p^{\pio}(x|x') \eta(o|x')$ and let $\eta^\pi$ denote $\eta$ marginalized over $d_\pi$: $\eta^\pi(o) = \sum_{x \in \X} d_\pi(x) \eta(o|x)$.
Then, we define $O$ as a random variable with $\eta^\pi$.
By decomposing this MI as $I(X_f;O| x_s) = H(X_f|x_s) - H(X_f|x_s, O)$, we can interprete this MI maximization as maximization of $H(X_f|x_s)$ and minimization of $H(X_f|x_s, O)$.
Maximizing $H(X_f|x_s)$ diversifies possible destinations of an agent.
Thus, we expect the resulting options to be diverse because choosing a different option is likely to lead to a different destination.
On the other hand, minimizing $H(X_f|x_s, O)$ makes the option state transition more deterministic, making the options more meaningful.

\begin{figure}[t]
  \centering
  \includegraphics[width=6.5cm]{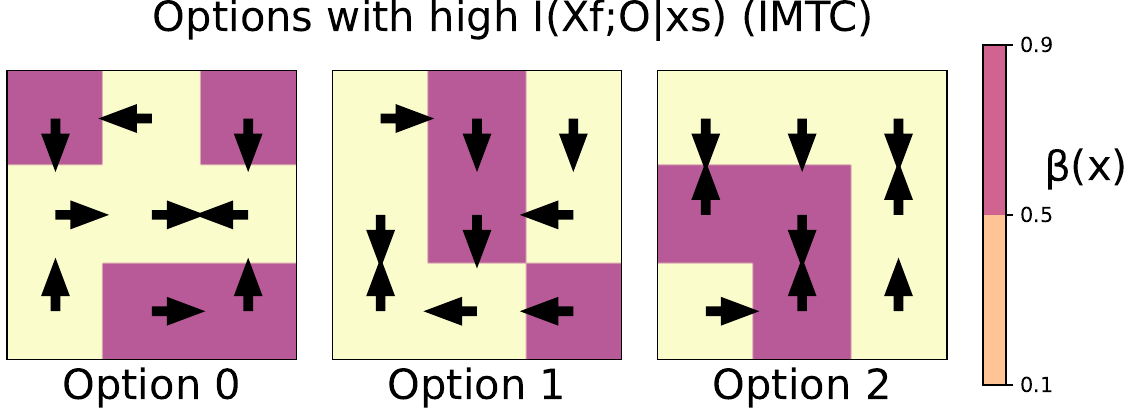}
  \includegraphics[width=6.5cm]{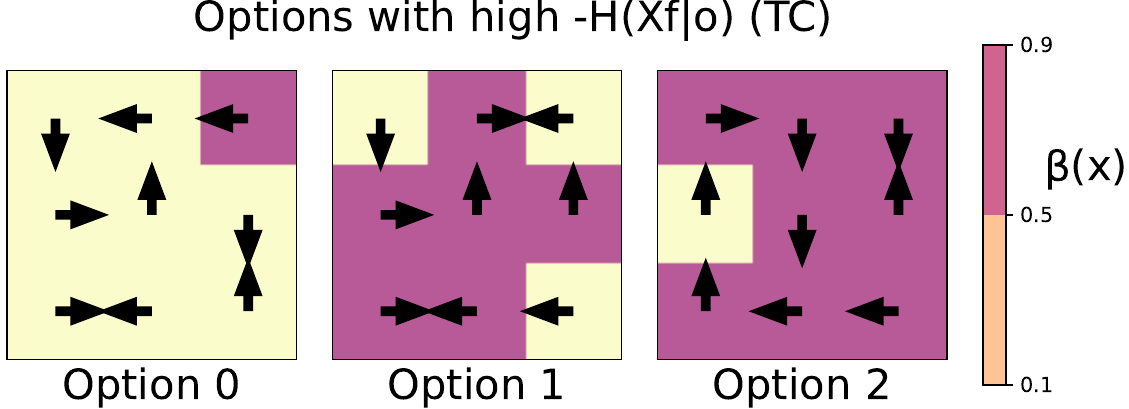}
  \vspace{-2mm}
  \caption{IMTC options (left) and TC options (right).}
  \vskip -0.1in
  \label{figure:random-searched-options}
\end{figure}

To build an intuition, we searched options that maximizes this MI by a simple blute-force method from a limited set of $\pio$ and $\bo$ in a $3 \times 3$ gridworld environment.
Specifically, we considered only deterministic intra-option polciies and $0.1$ or $0.8$ as the value of $\bo(x)$.
We set the number of options as $3$.
In this environment, an agent can take four actions: \texttt{Up}, \texttt{Down}, \texttt{Left}, and \texttt{Right}.
With probability $0.1$ the taken action fails and it takes an uniformly random action.
\Cref{figure:random-searched-options} shows the result.
Arrows represent intra-option policies and colors of the room represent termination probabilities.
Here, we can see the tendency that IMTC prefers options that have almost separated termination regions and different policies per option.
We also visualized the options that maximize $-H(X_f|o)$ (TC options) proposed by \citet{Harutyunyan2019TC}.
For TC options, we can see a similar tendency as IMTC options, but option $0$ and $1$ of TC share the intra-option policies on the left side of the room.
Thanks to the diversity term $H(X_f|x_s)$, IMTC successfully avoids such a policy overlap.
In addition, we can see that IMTC has a smaller termination regions than TC, which can enhance long-range abstraction.

We propose to maximize the MI \eqref{eq:infomax} by updating $\bo$ via gradient ascent w.r.t. $\theta_\beta$.
For this purpose, we now derive an unbiased estimator of the gradient.
First, we write the gradient of the MI using the option transition model $\po$ and marginalized option-transition model $P(x_f|x_s) = \sum_o \eta^\pi(o) \po (x_f|x_s)$.
\begin{proposition} \label{proposition:imoc-gradient-raw}
 Let $\bo$ be parameterized using a sigmoid function.
 Given a trajectory $\tau = x_s, \dots, x, \dots, x_f$ sampled by $\pio$ and $\bo$, we can obtain unbiased estimations of $\deriv H(X_f|x_s)$ and $\deriv H(X_f|x_s,O)$ by
 \begin{align}
  & \deriv H(X_f|x_s) = \E_{d^\pi, \eta} \left[
  - \deriv\ell_{\bo}(x) \bo(x) \Big(\log P(x|x_s) - \log P(x_f|x_s) \Big)
  \right] \label{eq:h-xf-xs-sampled} \\
  & \deriv H(X_f|x_s, O) = \E_{d^\pi, \eta} \left[
  - \deriv\ell_{\bo}(x) \bo(x) \Big(\log \po(x|x_s) - \log \po(x_f|x_s) \Big)
  \right] \label{eq:h-xf-xs-o-sampled},
 \end{align}
 where $\ell_\bo(x)$ denotes the logit of $\bo(x)$.
\end{proposition} \vspace*{-0.02in}
Note that the additional term $\bo$ is necessary because $x$ is not actually a terminating state.
The proof is based on Section 4 in \citet{Harutyunyan2019TC} and is given in \Cref{appendix:proofs:1}.

Now the targetting gradient can be written as:
\begin{align}
 & \deriv I(X_f; O| x_s) = \deriv H(X_f|x_s) - \deriv H(X_f|x_s, O) \notag \\
 &=\E_{d^\pi, \eta} \left[
 - \deriv\ell_{\bo}(x) \bo(x) \left(
   \log P(x|x_s) - \log P(x_f|x_s)
 - \left(\log \po(x|x_s) - \log \po(x_f|x_s)\right)
 \right) \right], \label{eq:mi-sampled}
\end{align}
which means that we can approximate MI maximization by estimating $\po$ and $P$.
However, in RL, learning a density model over the state space in RL is often difficult, especially with large or continuous state spaces.
For example, it has been tackled with data compression methods \citep{Bellemare2016PC} and generative models \citep{Ostrovski2017NDM}.
Hence, we reformulate the gradient using Bayes' rule to avoid estimating $\po$ and $P$.
The resulting term consists of the inverse option transition function $\pop(o|x_s, x_f)$, which denotes the probability of having an $o$ given a state transition $x_s, x_f$.
\begin{proposition} \label{proposition:imoc-gradient}
 We now have
  \begin{align} \label{eq:imoc-gradient}
    \deriv I(X_f; O|x_s)
      &= \deriv H(X_f|x_s) - \deriv H(X_f|x_s, O) \notag \\
      &= \E_{d^\pi, \eta} \left[ \deriv\ell_{\bo}(x) \bo(x) \Big(
           \log \pop(o|x_s, x) - \log \pop(o|x_s, x_f)
         \Big) \right].
  \end{align}
\end{proposition} \vspace*{-0.02in}
The proof is provided in \Cref{appendix:proofs:2}.
\Cref{eq:imoc-gradient} requires estimating of $\pop$ per updating $\pio$ and $\bo$, which is computational quite expensive.
Thus, we approximate approximate the gradient~\eqref{eq:imoc-gradient} by regressing a classification model over options $\ph(o|x_s, x_f)$ on sampled option transition
\cutsectionup
\section{Implementation} \label{section:implementation}
\cutsectiondown
Since IMTC can be combined with any on-policy RL methods, we choose PPO~\citep{Schulman2017PPO} as a base algorithm because of its stability and ease of implementation.
As notable implementation details, this section explains the estimation of $\pop$, advantage estimation, and our VIC implementation.
We provide further details and the full description of the algorithm in \Cref{appendix:impl-details}.

\cutsubsectionup
\subsection{Estimating $\pop$}  \label{subsection:implementation:pop}
\cutsubsectiondown
To estimate $\pop$, we employ a classification model over options $\ph(o|x_s, x_f)$ and regress it on sampled option transitions, as per \citet{Gregor2017VIC}.
However, our preliminary experiments observed that this online regression could be unstable because the supply of transition data depends on the termination probability and can drastically increase or decrease during training.
To address this problem, we maintain a replay buffer $B_\Op$, which stores option state transitions $\{(o, x_s, x_f)\}$ to stabilize the regression of $\ph$.
Note that using older option state transitions can introduce bias to $\ph$ because it depends on the current policy.
However, we found that this is not harmful when the capacity of the replay buffer is reasonably small.

\cutsubsectionup
\subsection{Advantage Estimation} \label{subsection:implementation:adv}
\cutsubsectiondown
The original PPO implementation employs GAE~\citep{Schulman2015GAE} for estimating the advantage, which is important for learning performance \citep{Andrychowicz2021LargeScale}.
Therefore, we employed two variants of GAE in experiments for option learning according to the experimental setup.
In the following paragraphs, we let $N$ denote the rollout length used for advantage estimation and let $t + k$ denote the time step at which the current option $o_t$ terminates.
Thus, we need to consider the effect of option-switching in advantage estimation When $k < N$.
Furthermore, we use two variants of the option-specific TD errors
$\delta(o_t) = R_t + \gamma \qo(x_{t + 1}, o_t) - \qo(x_t, o_t)$ and
$\delta_\textrm{U}(o_t) = R_t + \gamma \uo(x_{t + 1}, o_t) - \qo(x_t, o_t)$.

\cutparagraphup
\paragraph{Independent GAE for Reward-Free RL}
For no-reward experiments with VIC, we used the following variant of GAE:
\begin{align} \label{eq:independent-gae}
 \ah^o_{\textrm{ind}} = -\qo(x_t, o_t) +
 \sum_{i = 0}^{\min(k, N)} (\gamma \lambda)^i \delta(o_{t + i})
\end{align}
Here, we ignore the future rewards produced by other options after the current option $o_t$ terminates.
This formulation enhances learning diverse intra-option policies per option.

\cutparagraphup
\paragraph{Upgoing GAE for task adaptation}
For single-task learning, increasing the rollout step $N$ often speeds up learning \citep{Sutton2018RL}.
However, future rewards after option termination heavily depend on the selected option and have high variance, especially when learning diverse options.
This high variance of future rewards slows advantage learning and causes underestimation of $\ah^o$.
Thus, to prevent underestimation, we introduce an \textit{upgoing} GAE (UGAE) for estimating advantage with options:
\begin{align} \label{eq:upgoing-gae}
 \ah^o_{\textrm{upg}} = -\qo(x_t, o_t) +
 \begin{cases}
   \sum_{i = 0}^{k} (\gamma \lambda)^i \delta^o_{t + i}
   + \ubcomment{\max\left(
       \sum_{i=k + 1}^{N}  (\gamma \lambda)^i \delta(o_{t + i}), 0
     \right)}{upgoing estimation} & (k < N) \\
     \sum_{i = 0}^{N - 1} (\gamma \lambda)^i \delta^o_{t + i} +
     (\gamma \lambda)^N \delta_\textrm{U}(o_{t + N}) & (\mbox{otherwise}). \\
 \end{cases}
\end{align}
Like the upgoing policy update~\citep{Vinyals19AlphaStar}, the idea is optimistic regarding future rewards after option termination by taking the maximum over $0$.
We use $\ah^o_{\textrm{upg}}$ for task adaptation experiments in \Cref{section:experiments:transfer}, and confirmed its effectivity in the ablation study in \Cref{appendix:exp-details:vo}.

\cutsubsectionup
\subsection{VIC Implementation} \label{subsection:implementation:vic}
\cutsubsectiondown
In experiments, we show that IMTC helps an agent learn diverse options without reward signals.
For this purpose, we employ VIC \citep{Gregor2017VIC} as a method for providing intrinsic rewards.
Here we explain our VIC implementation.
VIC updates intra-option policies to maximize the lower bound of MI \eqref{eq:infomax} $H(O|x_s) - H(O| x_s, X_f) \geq H(O|x_s) + \E_{o, x_f}[\log q (o|x_s, x_f)]$ as rewards, where $q$ can be any distribution and called an \textit{option inference model}.
We use $\ph$ as $q$.
We also learn $\etah(o|xs)$ from sampled option transitions and approximate $H(O|x_s)$ by $H(O|x_s) = -\sum_{o\in\Op} \eta(o|x_s) \log \eta(o|x_s) = -\E\left[ \log \eta(o|x_s) \right] \approx \log \etah(o|x_s)$.
To this end, we giving an agent $R_\textrm{VIC} =  \cvic\left(\log \ph(o|x_s, x_f) - \log \etah(o|x_s)\right)$ as an immediate reward when an option $o$ terminates, where $\cvic$ is a scaling coefficient.
This is different from the original VIC implementation where $H(O|x_s)$ is treated as a constant.
However, we empirically found this formulation helps diversity options in our preliminary experiments.
\cutsectionup
\section{Experiments} \label{section:experiments}
\cutsectiondown
As presented in this section, we conducted two series of experiments to analyze the property of IMTC.
First, we qualitatively evaluated the diversity of options learned by IMTC in a reward-free setting.
We use VIC \citep{Gregor2017VIC} for intrinsic rewards in these experiments.
Second, we quantitatively test the reusability of learned options by task adaptation learned options on a specific task.
Through experiments, we compared our method with OC \citep{Bacon2017OC} and vanilla VIC with fixed $\bo$.
We used $\forall_x \bo(x) = 0.1$ for VIC since it performed the best in task adaptation tasks, while $0.05$ was used in \citet{Gregor2017VIC}.
In order to check only the effect of the different methods for learning beta $\bo$, the rest of the implementation is the same for these three methods.
That is, OC and vanilla VIC also use PPO and advantage estimation methods in \Cref{subsection:implementation:adv} for learning $\pio$.
In this section, we fix the number of options as $|\mathcal{O}| = 4$ for all option-learning methods.
We further investigated the effect of the number of options \Cref{appendix:exp-details}, where we confirmed that $|\mathcal{O}|=4$ is sufficient for most domains.
We did not compare IMTC with TC \citep{Harutyunyan2019TC} because our TC implementation failed to learn options with relatively small termination regions as reported in the paper, and there is no official public code for TC.
All environments that we used for experiments are implemented on the MuJoCo \citep{Todorov2012Mujoco} physics simulator.
We further describe the detail in \Cref{appendix:exp-details}.

\cutsubsectionup
\subsection{Intrinsic Option Learning with VIC} \label{section:experiments:intrinsic}
\cutsubsectiondown
We now compare the options learned by IMTC with options of other methods.
Generally, learned options depend on the reward structure in the environment.
Thus, to make sure that rewards are not designed intentionally to learn good options, we employ a reward-free RL setting where no reward is given to agents.
Instead, for all methods, we used VIC \citep{Gregor2017VIC} as intrinsic rewards for training each intra-option policy $\pio$.
We noted our implementation of VIC in \Cref{subsection:implementation:vic}.
We fix $\mu$ as $\mu(o|x) = \frac{1}{|\Op|}$ in this experiment.
Intra-option policies are trained by PPO \citep{Schulman2017PPO} and independent GAE \eqref{eq:independent-gae}.
We show network architectures and hyperparameters in \Cref{appendix:exp-details}.
We set the episode length to \num{1e4}, so an agent is reset to its starting position after \num{1e4} steps.
For all visualizations, we chose the best one from five independent runs with different random seeds.

\begin{figure}[t]
  \includegraphics[width=13.5cm]{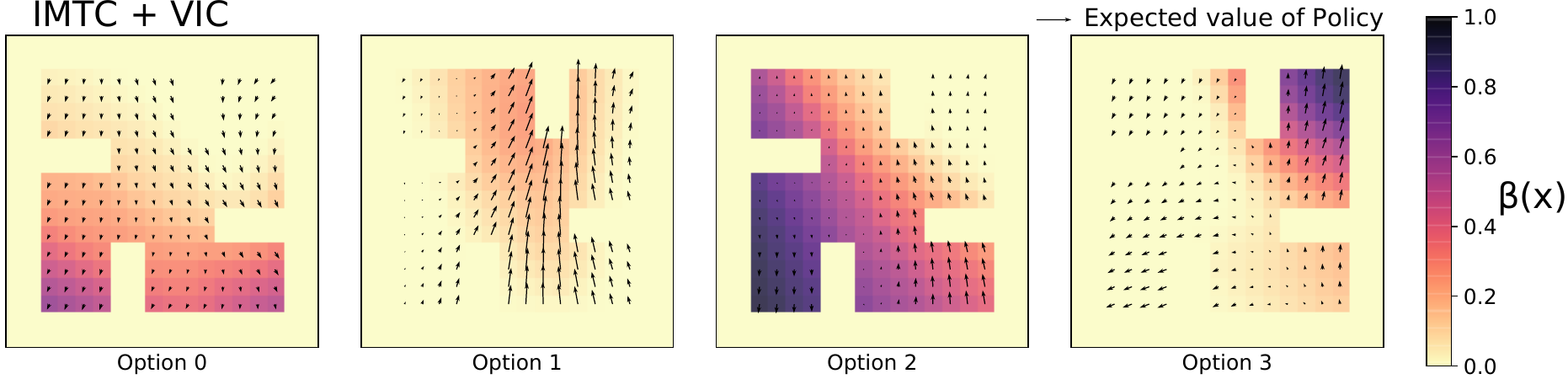}
  \includegraphics[width=13.5cm]{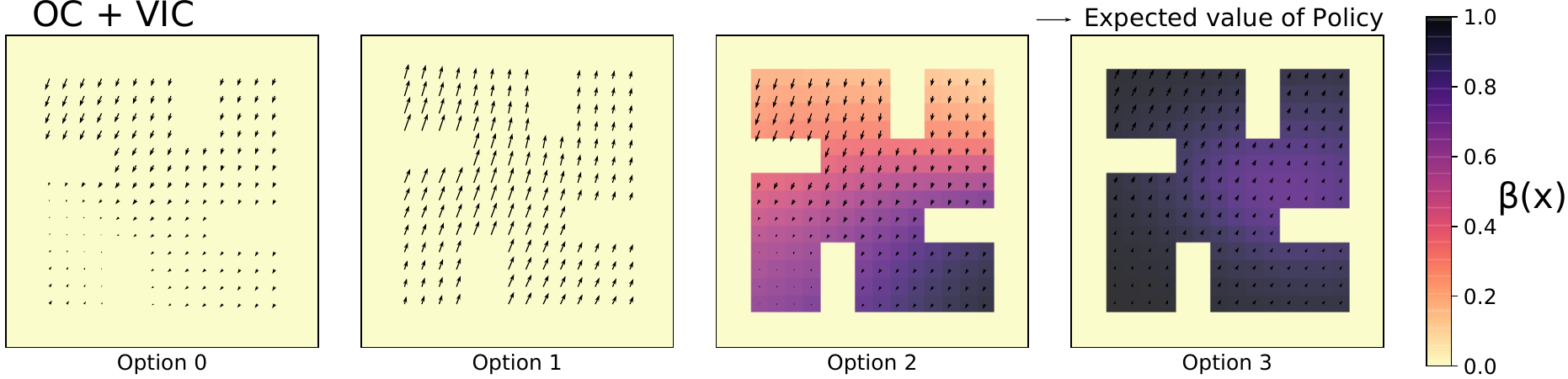}
  \includegraphics[width=13.5cm]{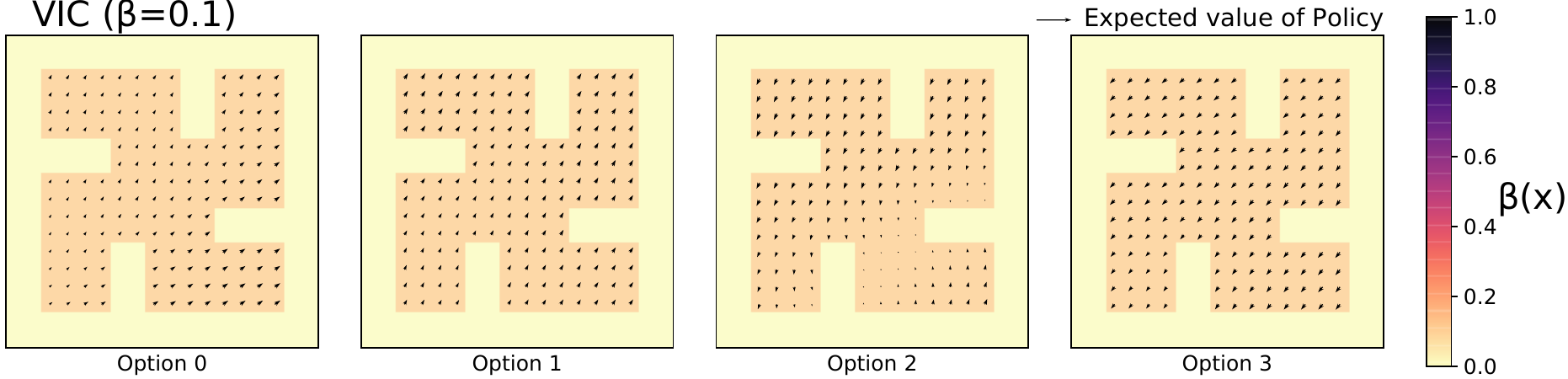}
  \vspace{-2mm}
  \caption{
    Learned intra-option policies ($\pio$) and termination probabilities ($\bo$) for each option in PointMaze after training \num{4e6} steps.
    Arrows show the expected value of action, and heatmaps show probabilities of each $\bo$.
    \textbf{First row:} Options learned by IMTC and VIC rewards.
    Intra-option policies have various directions, e.g., option $0$ points from lower left to lower right and option $1$ points from upper right to left.
    Also, termination regions are clearly separated across options.
    For example, option $2$ terminates at the lower left, and contrary, option $3$ terminates at the upper right.
    \textbf{Second row:} Options learned by OC and VIC rewards.
    Intra-option policies have two directions.
    Option $0$ and $2$ point to the lower left, and option $1$ and $3$ point to the upper right.
    Option $3$'s termination region is close to $1.0$ at all states, and thus, this option is rarely used.
    Contrary, termination probabilities of option $0$ and $1$ are close to $0$ at all states. Thus these options never terminate.
    \textbf{Third row:} Options learned by vanilla VIC with fixed $\bo$.
    Similar to OC, Intra-option policies have basically two directions.
    Option $0$ and $1$ point to the lower left, and option $2$ and $3$ point to the upper right.
  }
  \vskip -0.1in
  \label{figure:point-corridor-options}
\end{figure}

We visualized learned options in \texttt{PointReach} environment shown in \Cref{subfigure:pointreach}.
In this environment, an agent controls the ball initially placed at the center of the room.
The state space consists of positions $(\mathbf{x}, \mathbf{y})$ and velocities $(\Delta \mathbf{x}, \Delta \mathbf{y})$ of an agent, and the action space consists of acceralations $(\frac{\Delta \mathbf{x}}{\Delta \mathbf{t}}, \frac{\Delta \mathbf{y}}{\Delta \mathbf{t}})$.
\Cref{figure:point-corridor-options} shows the options learned in this environment after \num{4e6} steps.
Each arrow represents the mean value of intra-option policies, and the heatmaps represent $\bo$.
In this experiment, we observed the effect of IMTC clearly, for both termination regions and intra-option policies.
Termination regions are almost separated over options: option $0$ terminates at the lower, option $1$ at the upper-center, option $2$ at the left, and option $3$ at the upper right.
Only termination regions of option $0$ and $2$ overlap each other.
Contrary, OC failed to learn meaningful termination regions: option $0$ and $1$ never terminate, and option $3$ terminates almost everywhere.
Hence, we can say that IMTC certainly helps to learn diverse termination regions.
We also observed that the intra-option policies of IMTC are the most diverse.
While intra-option policies of OC and vanilla VIC point out roughly two directions (lower left and upper right), all intra-option policies of IMTC point in different directions.
Moreover, the magnitude of intra-option policies (shown by the length of arrows) is larger with IMTC.
This result suggests that IMTC diversifies and clarifies intra-option policies by diversifying termination regions.
\Cref{figure:point-billiard-options} also show options learned by IMTC and OC in \texttt{PointBilliard} domain, where we can see the same tendency.

\cutsubsectionup
\subsection{Transferring skills via task adaptation}
\cutsubsectiondown
\label{section:experiments:transfer}
Now we quantitatively test the reusability of learned options by task adaptation them with specific reward functions.
Specifically, we first trained agents with VIC rewards without external rewards as per the previous section.
Then we transferred agents to an environment with the same state and action space but with some external reward functions.
We prepared multiple reward functions, which we call tasks, for each domain and evaluated the averaged performance over tasks.
We compare IMTC with OC, vanilla VIC, and PPO without pre-training.
For a fair comparison, all of IMTC, OC, and vanilla use UGAE~\eqref{eq:upgoing-gae} and PPO-style policy updates for training intra-option policies.
We found UGAE is very effective for task adaptation and show the result in \Cref{appendix:exp-details:vo}.
For vanilla VIC, termination probability is fixed to $0.1$ through pre-training and task adaptation.
$\epsilon$-greedy based on $\qo$ with $\epsilon=0.1$ is used as the option selection policy $\mu$.
We hypothesize that diverse options learned by IMTC can help quickly adapt to given tasks, supposing the diversity of tasks.

\begin{figure}[t]
  \centering
  \begin{subfigure}[t]{3.2cm}
    \includegraphics[height=3cm]{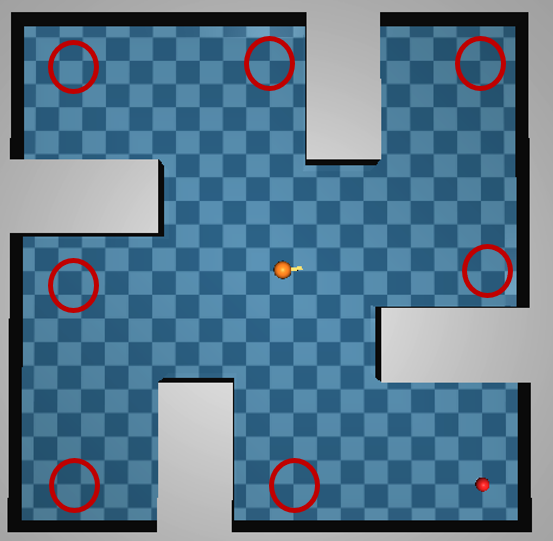}
    \caption{\texttt{PointReach}} \label{subfigure:pointreach}
  \end{subfigure}
  \begin{subfigure}[t]{3.2cm}
    \includegraphics[height=3cm]{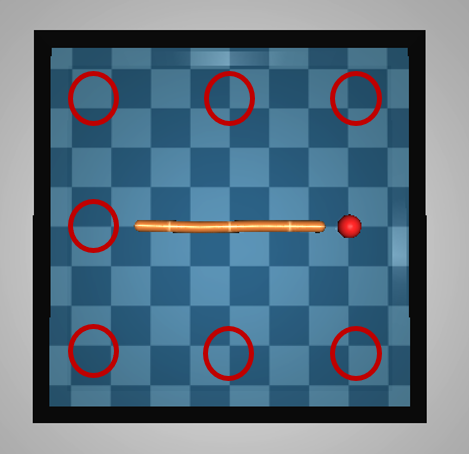}
    \caption{\texttt{SwimmerReach}}
  \end{subfigure}
  \begin{subfigure}[t]{3.2cm}
    \includegraphics[height=3cm]{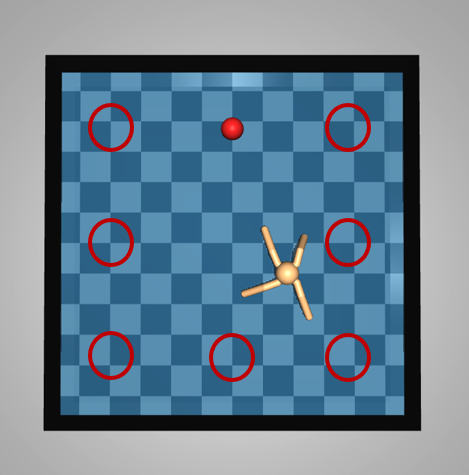}
    \caption{\texttt{AntReach}}
  \end{subfigure}
  \begin{subfigure}[t]{3.2cm}
    \includegraphics[height=3cm]{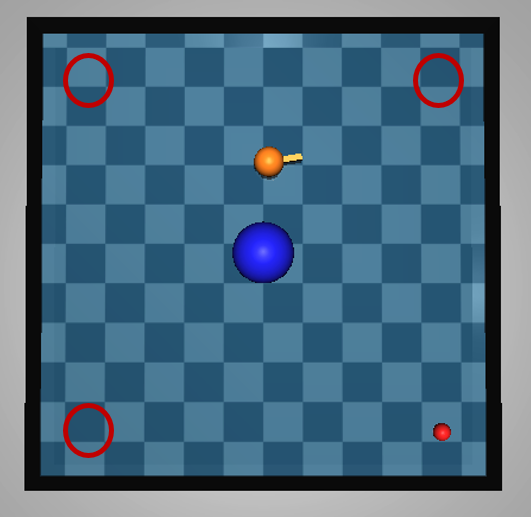}
    \caption{\texttt{PointBilliard}}
  \end{subfigure}
  \begin{subfigure}[t]{3.2cm}
    \includegraphics[height=3cm]{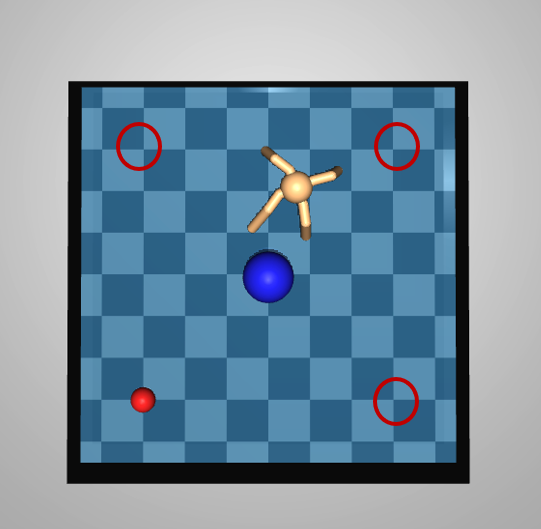}
    \caption{\texttt{AntBilliard}}
  \end{subfigure}
  \begin{subfigure}[t]{2.3cm}
    \includegraphics[height=3cm]{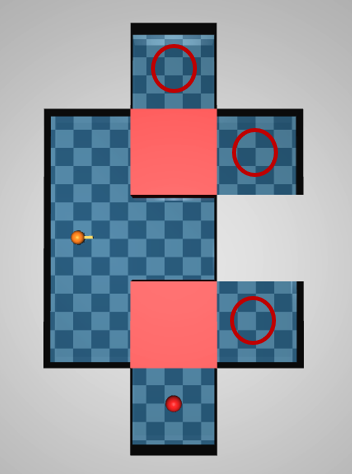}
    \caption{\texttt{PointPush}}
  \end{subfigure}
  \begin{subfigure}[t]{2.6cm}
    \includegraphics[height=3cm]{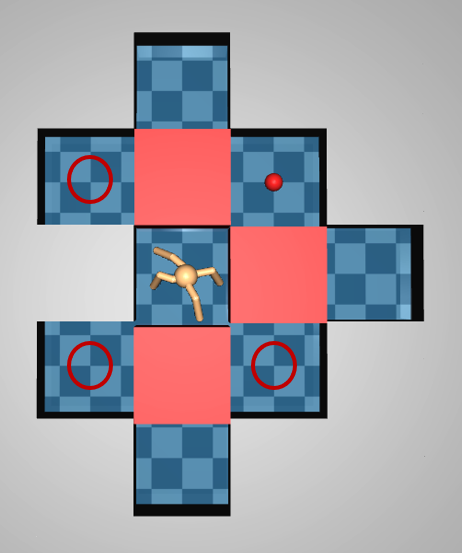}
    \caption{\texttt{AntPush}}
  \end{subfigure}
  \begin{subfigure}[t]{5.0cm}
    \includegraphics[width=5.0cm]{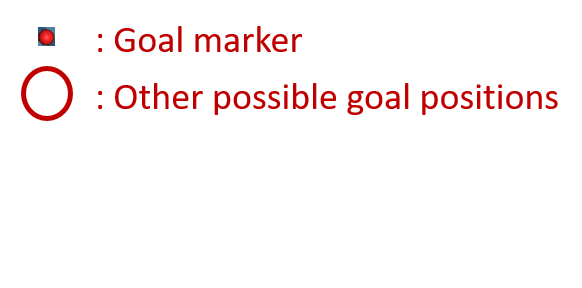}
  \end{subfigure}
  \vspace{-2mm}
  \caption{Domains used in task adaptation experiments.}
  \label{figure:transfer-tasks}
  \vskip -0.1in
\end{figure}

\begin{figure}[t]
  \centering
  \noindent
  \includegraphics[height=3.4cm]{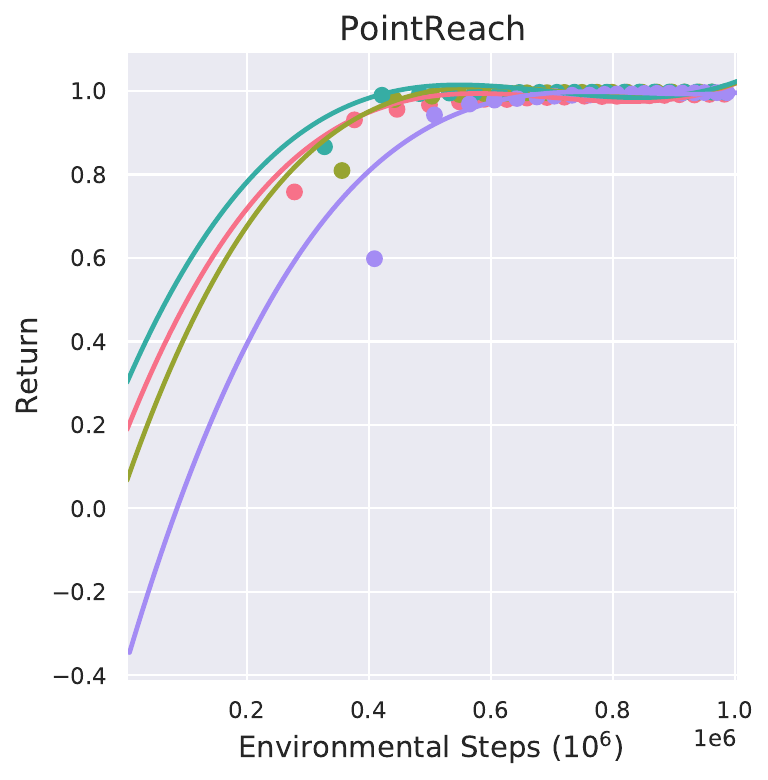}
  \includegraphics[height=3.4cm]{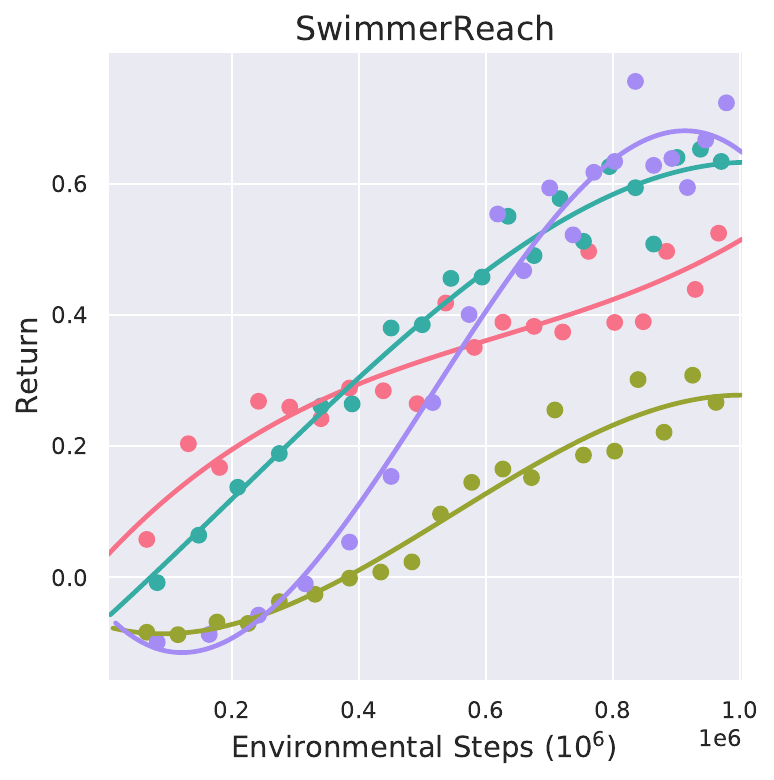}
  \includegraphics[height=3.4cm]{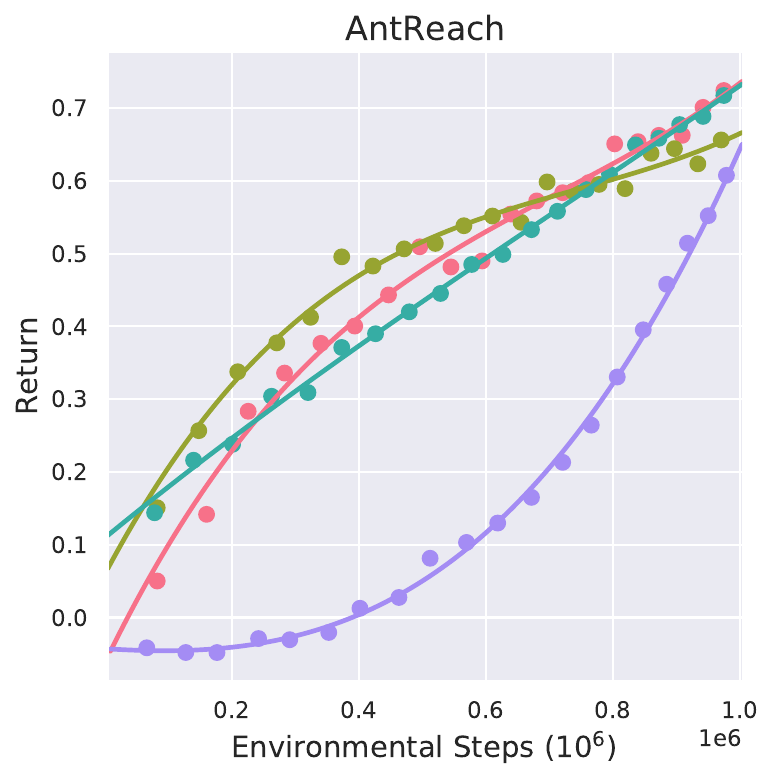}
  \includegraphics[height=3.4cm]{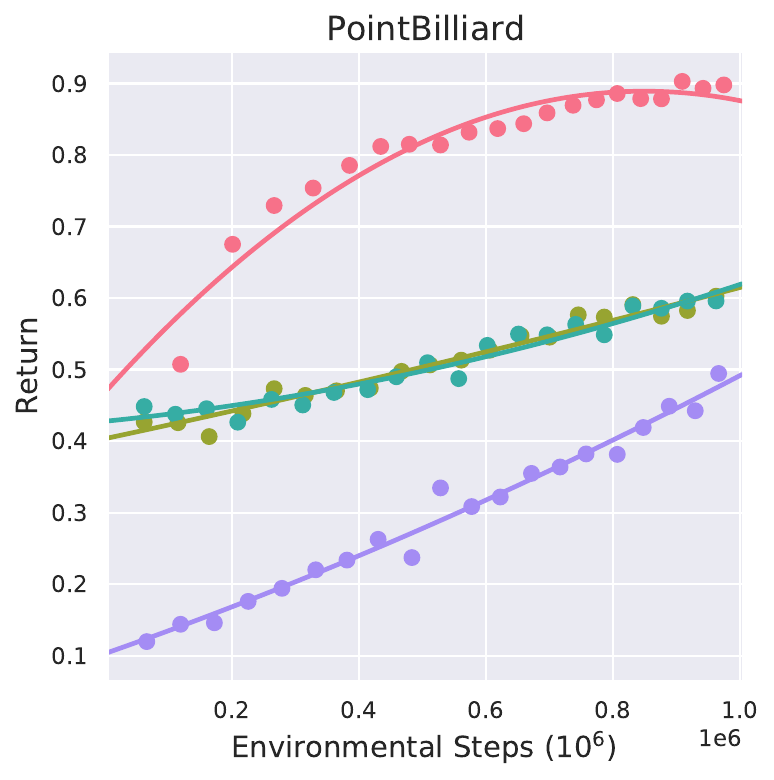}
  \includegraphics[height=3.4cm]{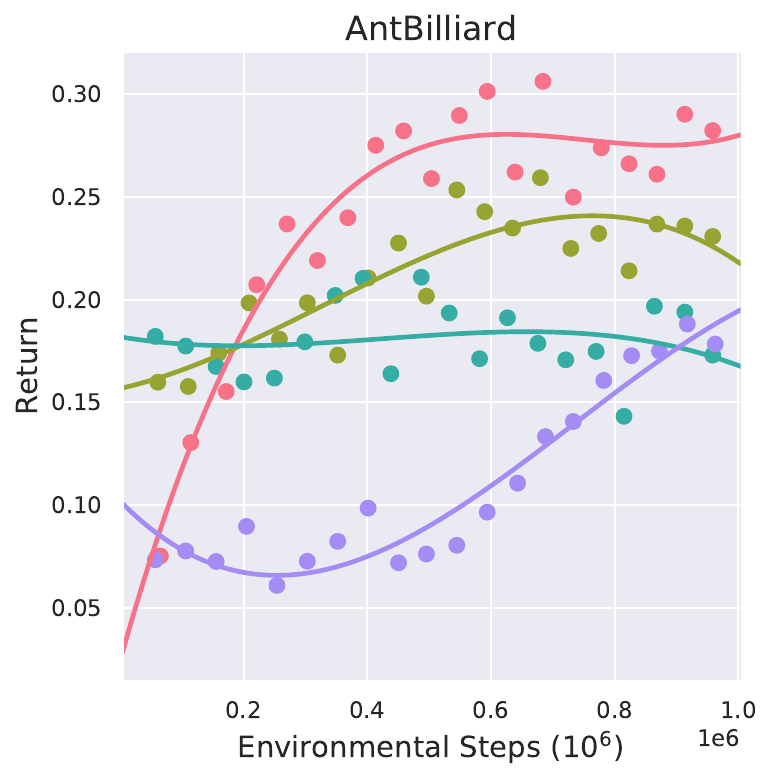}
  \includegraphics[height=3.4cm]{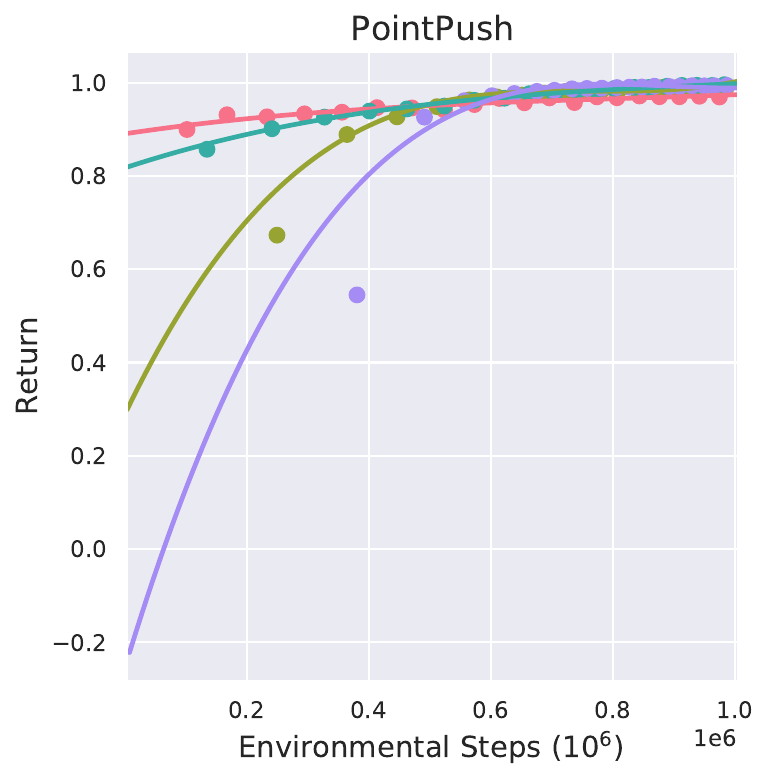}
  \includegraphics[height=3.4cm]{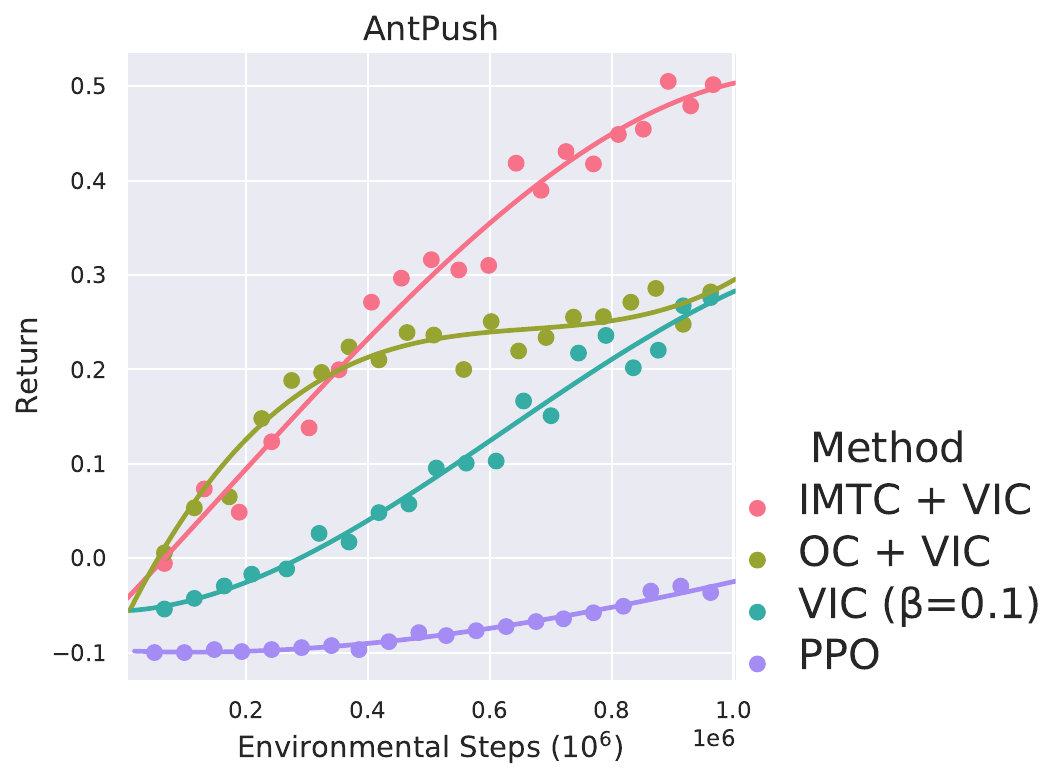}
  \vspace{-2mm}
  \caption{Learning curves for transfer learning experiments.}
  \label{figure:transfer-result}
  \vskip -0.1in
\end{figure}

\Cref{figure:transfer-tasks} shows all domains used for task adaptation experiments.
For simplicity, all tasks have goal-based sparse reward functions.
I.e., an agent receives $R_t=1.0$ when it satisfies a goal condition, and otherwise the control cost \num{-0.0001} is given.
Red circles show possible goal locations for each task.
When the agent fails to reach the goal after \num{1e3} steps, it is reset to a starting position.
\texttt{PointReach}, \texttt{SwimmerReach}, and \texttt{AntReach} are simple navigation tasks where an agent aim to just navigate itself to the goal.
We also prepared tasks with some complex tool use: in \texttt{PointBilliard} and \texttt{AntBilliard} an agent aims to kick the blue ball to the goal position, and in \texttt{PointPush} and \texttt{AntPush}, it has to push the block out of the way to the goal.
We pre-trained IMTC, OC, and vanilla VIC agents for \num{4e6} environmental steps and additionally trained them for \num{1e6} steps for each task.
\Cref{figure:transfer-result} shows learning curves and scatter plots drawn from five independent runs with different random seeds per domain.
\footnote{We used seaborn~\citep{Waskom2021Seaborn}'s \texttt{lmplot} with \texttt{order=3} method to draw plots.}
We observed that IMTC performed the best or was compatible with baselines except for \texttt{SwimmerReach}, where vanilla VIC and PPO finally outperformed IMTC.
Notably, IMTC outperforms other methods with a large margin in complex \texttt{PointBilliard}, \texttt{AntBilliard}, and \texttt{AntPush}.
This result suggests options learned by IMTC speed up adaptation to specific tasks, especially in complex domains where an agent needs to manipulate objects.
\cutsectionup
\section{Related Work} \label{section:related}
\cutsectiondown

\cutparagraphup
\paragraph{Discovering diverse options}
Since \citet{Sutton1999Option} first formulated the options, discovering good options has been challenging.
A classic concept for the goodness of options is a bottleneck region \citep{McGovern2001Bottleneck}, which refers to important states for reaching diverse areas, such as a passage between two rooms.
Once an important region is discovered, we can construct an option that guides an agent to that area.
Various approaches have been proposed to define bottleneck regions concisely and compute bottleneck options efficiently, including betweenness \citep{Simsek2008Betweenness}, Eigen options \citep{Machado2017Eigen}, and covering options \citep{Jinnai2019Cover}.
\citet{Barto2013BH} discussed the importance of diverse options for exploration, referring to an option construction method with a graph-based decomposition of an MDP~\citet{Vigorito2010Intrinsic}.
We share the same motivation with these methods but have a different approach.
While these methods construct \textit{point options} that bridge two states, we capture a set of states by learning $\bo$ directly, making it easy to scale up with function approximation.
To our knowledge, \citep{Jinnai2020DCO} only succeeded in scaling up point options to continuous options using approximated computation of the Laplacian.

\cutparagraphup
\paragraph{End-to-end learning of options}
As described in the previous paragraph, many studies have attempted to construct options and then train intra-option policies.
However, motivated by the recent success of RL with DNN, \citet{Bacon2017OC} proposed OC to train intra-option policies and termination functions in parallel using a neural network as a function approximator.
OC updates $\bo$ by gradient ascent, so that maximize $\qo$.
Thus, the resulting terminating regions heavily depend on the reward function.
OC also proposed a PG-style method for learning intra-option policies.
We used a similar method to OC for learning policies and values, but we employed a different method that does not depend on rewards for learning termination functions.
Also, we proposed UGAE \eqref{eq:upgoing-gae} for enhancing OC-style policy learning.
While OC maximizes the option-value function directly, many heuristic objectives have been proposed with similar architectures, including deliberation cost \citep{Harb2018Delib}, interest \citep{Khetarpal2020Interest}, and safety~\citep{Jain2018SafeOC}.
Notably, \citep{Harutyunyan2019TC} proposed the termination critic (TC) that maximizes an information-theoretic objective referred to as \textit{predictability} $-H(X_f|o)$.
Our method is inspired by TC and maximizes a similar information-theoretic objective for diversity rather than predictability.
In addition, TC requires estimating $\po (x_f|x_s)$ and a marginal distribution of $\po$, which can be quite difficult in environments with large or continuous state spaces.
While \citet{Harutyunyan2019TC} proposed to use approximated dynamic programming for estimating $\po$, it is even more difficult than estimating $Q^\pi$ or $V^\pi$ due to the lack of discounting and conditioning by $x_s$.
We avoid such difficult approximations using Bayes' rule, making IMTC more scalable.

\cutparagraphup
\paragraph{Mutual Information, Empowerment, and Skill Acquisition}
MI also appears in the literature regarding \textit{intrinsically motivated} RL~\citep{Singh2004IM}, as a driver of goal-directed behavior.
A well-known example is the \textit{empowerment} \citep{Klyubin2005Empowerment, Salge2014Empowerment}, obtained by maximizing MI between sequential $k$ actions and the resulting state
$I(a_t, ..., a_{t + k}; x_{t + k}|x_t) = H(x_{t + k} | x_t) - H(x_{t + k} | a_t, ..., a_{t + k}, x_t)$.
Empowerment represents both large degree of freedom and good preparedness: i.e., larger $H(x_{t + k} | x_t)$ implies that there can be more diverse future states, while smaller $ H(x_{t + k} | a_t, ..., a_{t + k}, x_t)$ indicates that an agent can realize its intention with greater certainty.
In RL literature, empowerment is often implemented by maximizing the variational lower bounds as intrinsic rewards~\citep{Mohamed2015VIM, Zhao2020EPC}.
We can interpret our objective $I(X_f;O| x_s)$ as a variant of empowerment where a fixed number of options represents action sequences.
\citet{Gregor2017VIC} also employed this interpretation and introduced VIC for training intra-option policies with fixed termination probabilities in a no reward RL setting.
Our experiments observed that IMTC helps VIC to learn meaningful intra-option policies.
As a variant of VIC, \citet{Baumli2021RVIC} proposed to use the reversed MI $I(X_s; O| x_f)$.
MI has been used for discovering diverse skills without termination functions.
\citet{Eysenbach2019DIAYN} proposed maximizing MI between skills and states $I(O; X)$, and \citet{Sharma2020DADS} extended the objective to a conditional MI $I(O; X'| x)$.
These methods also maximize variational lower bounds of MI as rewards.
\cutsectionup
\section{Conclusion} \label{section:conclusion}
\cutsectiondown
In this paper, we considered the problem of learning diverse options in RL.
To learn diverse termination regions of options in a scalable way, we proposed to maximize MI between options and terminating states per starting state.
We derived an unbiased gradient estimator to approximately maximize this MI, yielding the InfoMax Termination Critic~(IMTC) algorithm.
Also, we proposed a practical implementation of IMTC with enhanced advantage estimation in \Cref{subsection:implementation:adv}.
In reward-free experiments, we visualized that IMTC helped learn diverse and clear options combined with VIC.
We also showed that options learned by IMTC help an agent to quickly adapt to a specific reward function by transferring learned options.

Although our experiments observed that IMTC can learn clear and meaningful options, a problem is that learning of a classification model $\ph$ heavily depends on exploration.
For example, an agent cannot explore a room well, it would not be able to learn sufficiently diverse options.
Investigating the relationship between diversity and exploration would be an interesting approach.
Also, our analysis in \Cref{figure:random-searched-options} suggests that IMTC options can fall into small loops, forming uninteresting options.
To prevent this, considering distances between states, e.g., by using bisimulation metric \citep{Castro2010Bisimulation}, is a plausible research direction.
Another limitation of our method is that it requires on-policy data for training, making it difficult to use with off-policy RL methods.
Thus, updating termination regions in an off-policy way is an interesting future challenge.
\cutsectionup
\section*{Reproducibility statement}
\cutsectiondown
We publish anonymized source code used for all our experiments on \url{https://anonymous.4open.science/r/imtc-anonymized-code-E5D1/}.

\bibliography{references}

\clearpage
\appendix
\section{Omitted Proofs} \label{appendix:proofs}

\subsection{Proof of \Cref{proposition:imoc-gradient-raw}}
\label{appendix:proofs:1}
First, we make an assumption on the dependence of $\eta$ from $\beta$.
\begin{assumption} \label{assumption:independence}
  The empirical distribution of options $\eta(o|x_s)$ is independent of termination conditions $\cup_{o \in \Op} \bo$.
\end{assumption}

This assumption does not strictly hold, and we can consider using, e.g., two-time scale optimization to suppress the distribution shift of $\eta$ by updating $\bo$.
We employed a replay buffer to mitigate this issue in \Cref{subsection:implementation:pop}.

\begin{lemma} \label{lemma:1}
  The following equations hold.
  \begin{align}
    \deriv H(X_f|x_s) = - \sum_{o} \eta(o|x_s)
      \sum_{x}&\po(x|x_s) \deriv\ell_{\bo}(x) \notag \\
      &\Big[
        \log P(x|x_s) + 1
          - \sum_{x_f} \po(x_f|x) \Big( \log P(x_f|x_s) + 1 \Big)
       \Big] \label{eq:h-xf-xs} \\
    \deriv H(X_f|x_s, O) = - \sum_{o} \eta(o|x_s)
      \sum_{x}&\po(x|x_s) \deriv\ell_{\bo}(x) \notag \\
      &\Big[
        \log \po(x|x_s) + 1
          - \sum_{x_f} \po(x_f|x) \Big( \log \po(x_f|x_s) + 1 \Big)
       \Big] \label{eq:h-xf-xs-o}
  \end{align}
\end{lemma}

Then, sampling $x, x_f$ from $d^\pi$ and $o$ from $\eta$,
\begin{align}
 & \deriv H(X_f|x_s) = \E_{d^\pi, \eta} \left[
 - \deriv\ell_{\bo}(x) \bo(x) \Big(\log P(x|x_s) - \log P(x_f|x_s) \Big)
 \right] \label{eq:ap1} \\
 & \deriv H(X_f|x_s, O) = \E_{d^\pi, \eta} \left[
 - \deriv\ell_{\bo}(x) \bo(x) \Big(\log \po(x|x_s) - \log \po(x_f|x_s) \Big)
 \right] \label{eq:ap2}.
\end{align}

\paragraph{Proof of \Cref{lemma:1}}
\begin{proof}
  First, we prove \Cref{eq:h-xf-xs}.
  We have:
  \begin{align*}
    & \deriv H(X_f|x_s) = -\deriv \sum_{x_f} P(x_f|x_s) \log P (x_f|x_s) \\
    & = -\sum_{x_f} \Big(\deriv P(x_f|x_s) \log P(x_f|x_s)
    + P(x_f|x_s) \frac{\deriv P (x_f|x_s)}{P(x_f|x_s)} \Big) \\
    & = -\sum_{x_f} \deriv P(x_f|x_s) \Big( \log P(x_f|x_s) + 1 \Big) \\
    & = -\sum_{x_f} \sum_o \eta(o|x_s)
         \ubcomment{\deriv \po(x_f|x_s)}{Apply \cref{eq:termination-gradient}}
         \Big( \log P(x_f|x_s) + 1 \Big) \\
    & = -\sum_{x_f} \sum_o \eta(o|x_s)
         \sum_{x} \po(x|x_s) \deriv \ell_{\bo}(x)(\indic{x_f=x} - \po(x_f|x))
         \Big( \log P(x_f|x_s) + 1 \Big) \\
    & = -\sum_o \eta(o|x_s) \sum_{x} \po(x|x_s) \deriv \ell_{\bo}(x)
         \sum_{x_f}(\indic{x_f=x} - \po(x_f|x))
         \Big( \log P(x_f|x_s) + 1 \Big) \\
    & = -\sample{\sum_o \eta(o|x_s)}
         \sample{\sum_{x}\po(x|x_s)}
         \deriv \ell_{\bo}(x)
         \times \Big[
         \log P(x|x_s) + 1 - \sample{\sum_{x_f} \po(x_f|x)}
         \Big( \log P(x_f|x_s) + 1 \Big)
         \Big]. \\
  \end{align*}
  Sampling $x, x_f, o$, we get \eqref{eq:ap1}.
  Then we prove \Cref{eq:h-xf-xs-o}.
  \begin{align*}
    & \deriv H(X_f|x_s, O) = -\deriv \sum_{o} \eta(o|x_s) \sum_{x_f} \po(x_f|x_s) \log \po(x_f|x_s) \\
    & = -\sum_{o} \eta(o|x_s) \sum_{x_f} \Big(\deriv\po(x_f|x_s) \log \po(x_f|x_s)
        + \po(x_f|x_s) \frac{\deriv\po(x_f|x_s)}{\po(x_f|x_s)} \Big) \\
    & = -\sum_{o} \eta(o|x_s) \sum_{x_f}
         \ubcomment{\deriv \po(x_f|x_s)}{Apply \cref{eq:termination-gradient}}
         \Big( \log \po(x_f|x_s) + 1 \Big) \\
    & = -\sum_{o} \eta(o|x_s) \sum_{x_f} \sum_{x} \po(x|x_s) \deriv \ell_{\bo}(x)(\indic{x_f=x} - \po(x_f|x))
         \Big( \log \po(x_f|x_s) + 1 \Big) \\
    & = -\sum_{o} \eta(o|x_s) \sum_{x} \po(x|x_s) \deriv \ell_{\bo}(x)\sum_{x_f}(\indic{x_f=x} - \po(x_f|x))
         \Big( \log \po(x_f|x_s) + 1 \Big) \\
    & = -\sample{\sum_{o} \eta(o|x_s)}
         \sample{\sum_{x} \po(x|x_s)}
         \deriv \ell_{\bo}(x)
         \times \Big[ \log \po(x|x_s) + 1 -
         \sample{\sum_{x_f}\po(x_f|x)} \Big( \log \po(x_f|x_s) + 1 \Big)\Big] \\
  \end{align*}
  Sampling $x, x_f, o$, we get \cref{eq:ap2}.
\end{proof}

\subsection{Proof of \Cref{proposition:imoc-gradient}}
\label{appendix:proofs:2}
\begin{proof}
  First, using Bayes' rule, we have:
  \begin{align*}
    \po (x_f|x_s) = \frac{\Pr(o|x_f, x_s)\Pr(x_f|x_s)}{\Pr(o|x_s)} = \frac{\pop(o|x_f, x_s)P(x_f|x_s)}{\eta(o|x_s)}
  \end{align*}
  Then, we have:
  \begin{align*}
    \log \po (x_f|x_s) - \log P (x_f|xs)
    &= \log \frac{\po(x_f|x_s)}{P(x_f|xs)} \\
    &= \log \frac{\frac{\pop(o|x_s, x_f)P(x_f|x_s)}{\eta(o|x_s)}}{P(x_f|xs)} \\
    &= \log \frac{\pop(o|x_s, x_f)}{\eta(o|x_s)}
  \end{align*}
  Applying this equation to the \cref{eq:mi-sampled}, we have:
  \begin{align*}
    \deriv I(X_f; O| x_s)
    &= \deriv H(X_f| x_s) - \deriv H(X_f|x_s, O) \notag \\
    &= \E_{d^\pi, \eta} \left[
       - \deriv\ell_{\bo}(x) \bo(x) \Big(
         \log P(x|x_s) - \log P(x_f|x_s)
         - \log \po(x|x_s) + \log \po(x_f|x_s)
       \Big) \right] \notag \\
    &= \E_{d^\pi, \eta} \left[ \deriv\ell_{\bo}(x) \Big(
         \Big( \log \po(x|x_s) - \log P(x|x_s) \Big) -
         \Big( \log \po(x_f|x_s) - \log P(x_f|x_s) \Big)
       \Big) \right] \notag \\
    &= \E_{d^\pi, \eta}  \left[ \deriv\ell_{\bo}(x) \Big(
         \log \frac{\pop(o|x_s, x)}{\eta(o|x_s)} -
         \log \frac{\pop(o|x_s, x_f)}{\eta(o|x_s)}
       \Big) \right] \notag \\
    &= \E_{d^\pi, \eta} \left[ \deriv\ell_{\bo}(x) \Big(
         \log \pop(o|x_s, x) - \log \pop(o|x_s, x_f)
       \Big) \right] \\
  \end{align*}
\end{proof}
\section{Implementation Details} \label{appendix:impl-details}

\paragraph{Clipped $\beta$ loss}
Common PPO implementation updates $\pi_\theta$ multiple times.
However, our preliminary experiments observed that performing multiple updates for $\bo$ led to destructively large updates and resulted in the saturation of $\bo$ to zero or one.
Hence, to perform PPO-style multiple updates, we introduce a clipped objective of \cref{eq:imoc-gradient}:
\begin{align} \label{eq:clipped-beta}
  L^\textrm{CLIP}(\theta_\beta) =
  \textrm{clip}(
    \ell_{\bo}(x)
    - \ell_{\bo_{\textrm{old}}}(x),
    -\epsilon_{\beta}, \epsilon_{\beta}
  )
  \bo_{\textrm{old}}(x) \Big(
    \log \pop(o|x_s, x) - \log \pop(o|x_s, x_f)
  \Big),
\end{align}
where $\epsilon_{\beta}$ is a small coefficient, and $\bo_{\textrm{old}}$ is an old $\bo$ before updating.
We also add maximization of the entropy of $\bo$ for preventing the termination probability saturating on zero or one.
To this end, we maximize $L^\textrm{CLIP}(\theta_\beta) + \centb H(\bo(x))$ via gradient ascent w.r.t. $\theta_\beta$, where $\centb$ is a weight of the entropy bonus.

\paragraph{Full description of the algorithm}

\begin{algorithm}[t]
  \caption{InfoMax Termination Critic with VIC, PPO Style}
  \label{algorithm:imtc-ppo}
  \begin{algorithmic}[1]
  \State{\textbf{Given:} Initial option-value $\qo$, option-policy $\pio$, and termination function $\bo$}.
  \State{Let $B_\Op$ be a replay buffer for storing option-transitions.}
  \For{$k = 1, ...$}
    \For{$i = 1, 2, ..., N$} \Comment{Collect experiences from environment}
      \State{Sample termination variable $b_i$ from $\beta^{o_{i}}(x_i)$}
      \If{$b_i = 1$}
        \State{
          Store ($x_s, x_f, o_{i}$), ($x_{s + 1}, x_f, o_{i}$), ..., ($x_{s + h}, x_f, o_{i}$)
          to the replay buffer $B_\Op$
        } \label{algorithm:store-transitions}
      \EndIf
      \State{Choose next option $o_{i + 1}$ by $\epsilon$-Greedy}
      \State{Receive reward $R_i$ and state $x_{i + 1}$, taking $a_i \sim \pi_{o_{i + 1}}(x_i)$}
    \EndFor
    \For{$k = 1, 2, ...,$ Num. of PPO epochs} \Comment{Optimize $\pio, \qo$, and $\bo$}
      \ForAll{$x_i$ in the trajectory} \label{algorithm:minibatch-opt}
        \State{Compute $\rvic$ from the target network}
        \State{Compute $\ah^o_{\textrm{ind}}$ by \eqref{eq:independent-gae} using $\rvic$}
        \State{Update $\pi^o(a_i|x_i)$ via PPO using $\ah^o$}
        \State{Update $\qo(x_i, o)$ to regress to $\ah^o$}
        \If{$o_i$ has already terminated}
          \State{Update $\bo(x_i)$ via \eqref{eq:imoc-gradient}}
        \EndIf
      \EndFor
    \EndFor
    \State{Train $\hat{p}$ and $\hat{\mu}$ by option-transitions sampled from $B_\Op$}
    \If{$k \mod \kvic$}
      \State{Update the target network for VIC}
    \EndIf \label{algorithm:update-vic-net}
  \EndFor
  \end{algorithmic}
\end{algorithm}

\Cref{algorithm:imtc-ppo} shows a full description of our implementation of IMTC on PPO when combined with VIC rewards.
As of the original PPO, it is built on the A2C-style~\citep{Mnih2016A3C, Wu2017ACKTR} architecture with multiple synchronous actors and a single learner.
First, we collect $N$-step experiences interacting with environments.
At \cref{algorithm:store-transitions}, we append tuples corresponding to option transitions $(x_s, x_f, o_{i}), ..., (x_{s + h}, x_f, o_{i})$ to $B_\Op$.
Here, we do not use all transitions and store first $h_\textrm{max}$ options to prevent memory shortage.
We used $h_\textrm{max}=10$ or $h_\textrm{max}=20$.
Then we update $\pio$, $\qo$, and $\bo$.
\Cref{algorithm:minibatch-opt} is done via minibatch sampling in the actual implementation.
We also update $\ph$ for estimating the gradient~\eqref{eq:imoc-gradient}, sampling from the replay buffer $B_\Op$.
We empirically found that rapidly changing $\rvic$ leads to unstable learning, especially when IMTC is used in parallel.
Thus, we employ a target network to compute $\rvic$ and periodically update it at \cref{algorithm:update-vic-net}.
We used $\kvic=10$ or $\kvic=20$ in the experiments.
\section{Experimental Details} \label{appendix:exp-details}

Our anonymized code used for all our experiments is on \url{https://anonymous.4open.science/r/imtc-anonymized-code-E5D1/}.

\subsection{Network Architecture} \label{appendix:exp-details:nn}
\begin{figure}[t]
  \centering
  \includegraphics[width=9cm]{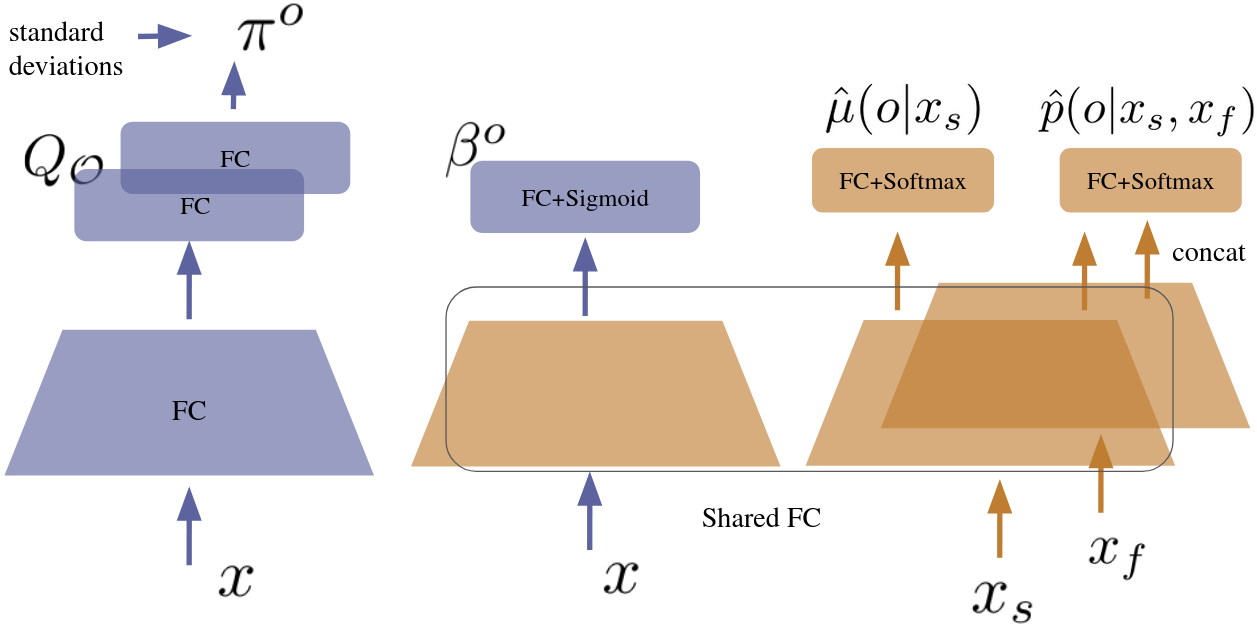}
  \caption{
      Neural Network architecture used
  }
  \label{figure:nn}
\end{figure}
\Cref{figure:nn} illustrates the neural network architecture used in our experiments.

In continuous control experiments, we encode the state by two fully connected layers with 64 units.
$\pio$ is parameterized as a Gaussian distribution with separated networks for standard derivations per option, similar to \citet{Schulman2015TRPO}.
We used ReLU as an activator for all hidden layers and initialized networks by the orthogonal~\citep{Saxe2013Exact} initialization in all experiments.
Note that the last layer for intra-option policy $\pio$ was initialized with small values, following the standard practice~\citep{Andrychowicz2021LargeScale}.
We used the Adam~\citep{Kingma2014Adam} optimizer in all experiments.
Unless otherwise noted, we used the default parameters in PyTorch~\citep{Paszke2019PyTorch} 1.8.1.

\subsection{Hyperparameters}

\begin{table}
  \centering
  \begin{tabular}{c|c}
    \hline
    Description & Value \\ [0.5ex]
    \hline\hline
    $\gamma$ & 0.99 \\
    \hline
    Adam Learning Rate & \num{3e-4} \\
    \hline
    Adam $\epsilon$ & \num{1e-4} \\
    \hline
    Clip parameter for $\ell_{\bo}$ ($\epsilon_\beta$) & 0.05 \\
    \hline
    Num. timesteps per rollout & 256 \\
    \hline
    Num. actors & 16 \\
    \hline
    GAE $\lambda$ & 0.95 \\
    \hline
    Num. epochs for PPO & 10 \\
    \hline
    Minibatch size for PPO & 1024 \\
    \hline
    Weight of $H(\pio)$ ($\cent$) & 0.001 \\
    \hline
    Weight of $H(\bo)$ ($\centb$) & 0.01 \\
    \hline
    Gradient clipping & 0.5 \\
    \hline
    Capacity of $B_\Op$ & 8192 \\
    \hline
    Max num. option transitions to store ($h_\textrm{max}$) & 20 \\
    \hline
    Num. epochs for training $\hat{p}$ and $\hat{\mu}$ & 4 \\
    \hline
    Minibatch size for training $\hat{p}$ and $\hat{\mu}$ & 2048 \\
    \hline
    Scaling of $\rvic$ ($\cvic$) & 0.005 \\
    \hline
    Synchronizing interval of the target network for VIC ($\kvic$) & 20 \\
    \hline
  \end{tabular}
  \vspace{0.1in}
  \caption{Hyperparameters used in continuous control experiments}
  \label{table:imtc-params-mujoco}
\end{table}


\Cref{table:imtc-params-mujoco} shows all hyperparameters used in IMTC + VIC experiments on MuJoCo continuous control tasks.

\subsection{Environment Implementation}
\label{appendix:exp-datails:environment}
Gridworld is based on RLPy (\citet{Geramifard2015RLPy}, BSD3 License).
We constructed continuous control environments on the MuJoCo (commercial license, \citet{Todorov2012Mujoco}), using OpenAI Gym (MIT license, \citet{Brockman2016Gym}).
Especially, point environments are implemented based on ``PointMaze'' in rllab (MIT license, \citet{Duan2016rllab}) with some modifications, mainly around collision detection.
We also refered to the modified PointMaze code\footnote{\url{https://github.com/tensorflow/models/tree/v2.3.0/research/efficient-hrl}} (Apache 2.0 license) relased by \citet{Nachum2018HIRO}.
The swimmer robot is originally used in \citet{Coulom2002Swimmer}.

\subsection{Computational Resources}
\label{appendix:exp-datails:resources}
All experiments are conducted on a private cluster with NVIDIA V100 and P100 GPUs.
On the cluster, training IMTC with VIC on MuJoCo PointMaze domain for \num{4e6} steps takes about 27 minutes.

\subsection{Effectivity of UGAE} \label{appendix:exp-details:vo}
\begin{figure}[t]
  \centering
  \noindent
  \includegraphics[height=3.4cm]{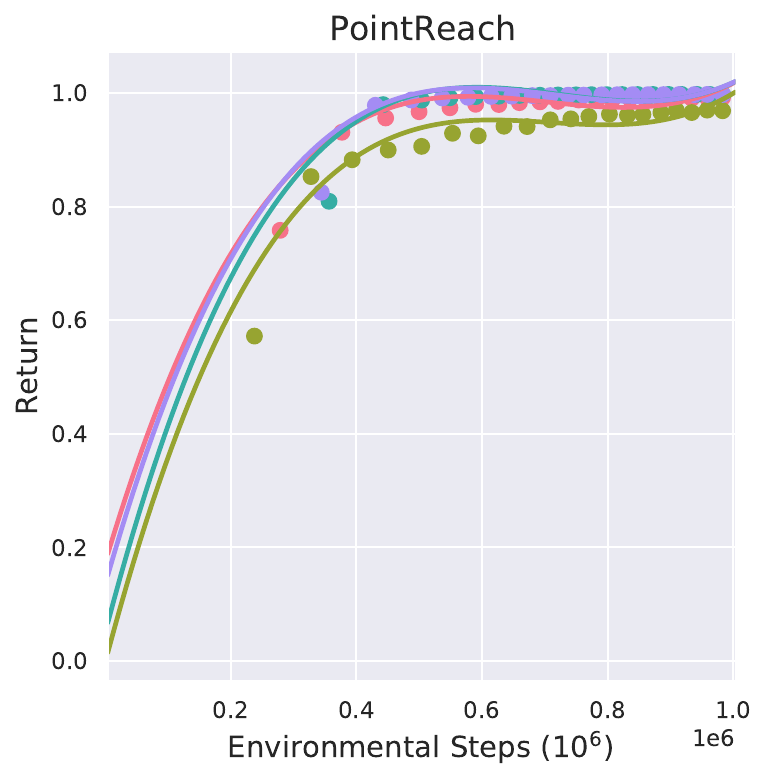}
  \includegraphics[height=3.4cm]{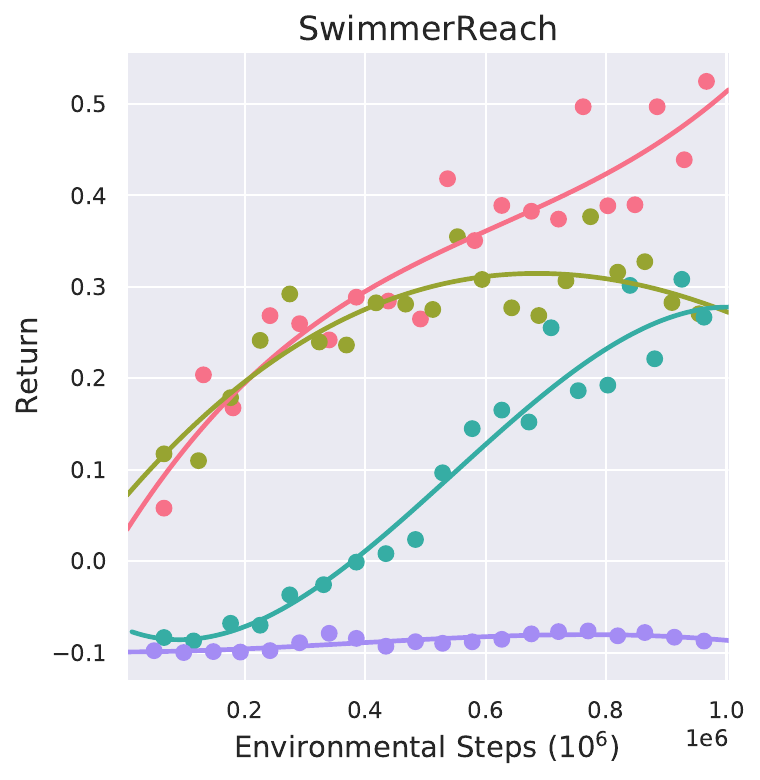}
  \includegraphics[height=3.4cm]{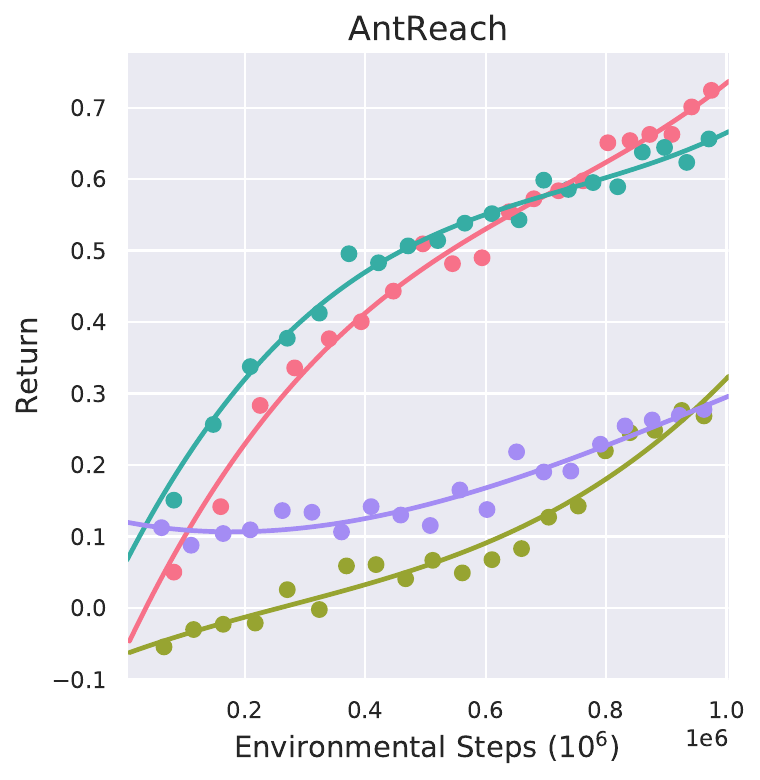}
  \includegraphics[height=3.4cm]{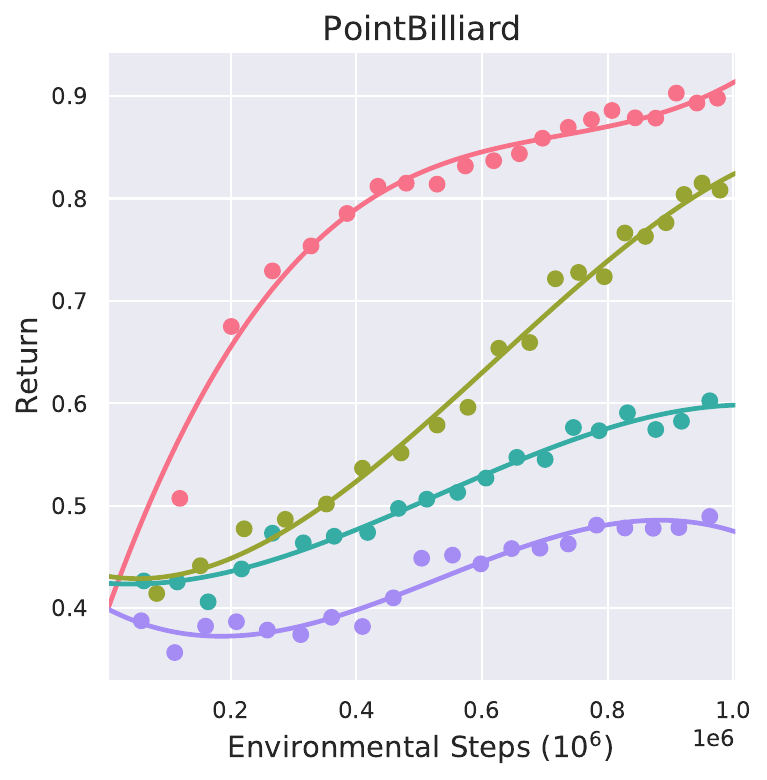}
  \includegraphics[height=3.4cm]{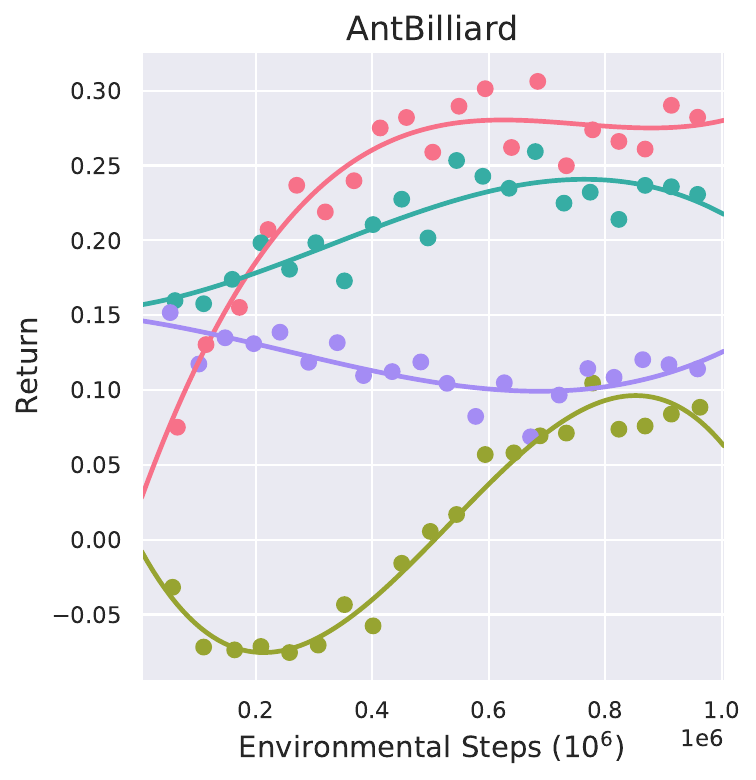}
  \includegraphics[height=3.4cm]{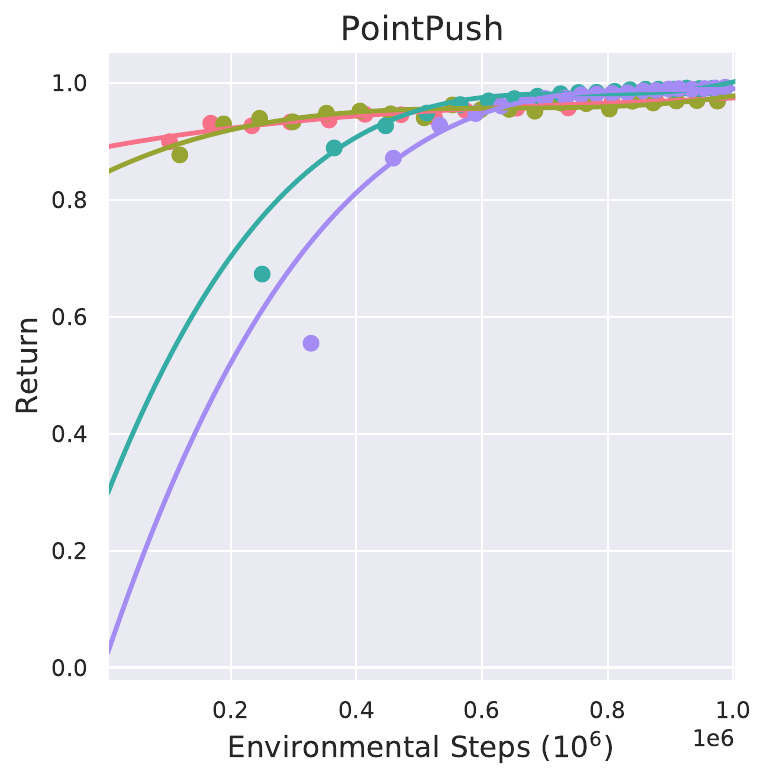}
  \includegraphics[height=3.4cm]{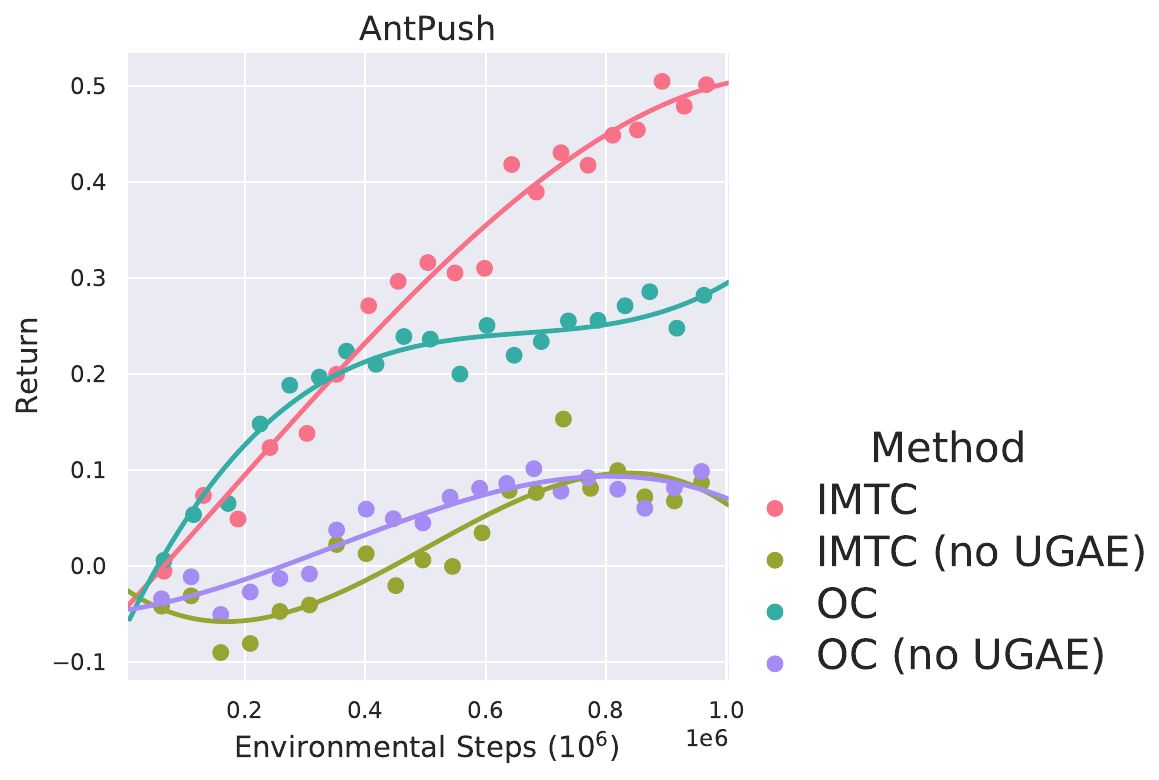}
  \vspace{-2mm}
  \caption{Learning curves for transfer learning experiments with }
  \label{figure:vo-result}
  \vskip -0.1in
\end{figure}
To investigate the effect of UGAE \eqref{eq:upgoing-gae}, we compared IMTC and OC with and without UGAE in task adaptation experiments.
\texttt{IMTC (no UGAE)} and \texttt{OC (no UGAE)} estimate the adavantage ignoring the future rewards after switching options, similar to advantage estimation proposed by \citet{Bacon2017OC}.
\Cref{figure:vo-result} shows the result.
We can see that UGAE improves the performance in all domains for both IMTC and OC.

\subsection{Number of Options} \label{appendix:exp-details:nopt}
\begin{figure}[t]
  \centering
  \noindent
  \includegraphics[height=3.4cm]{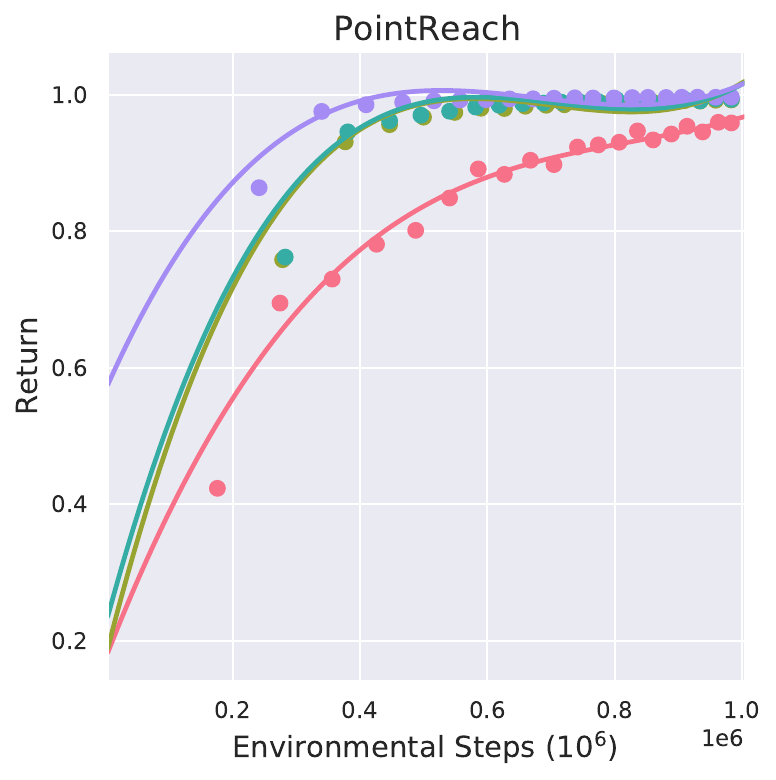}
  \includegraphics[height=3.4cm]{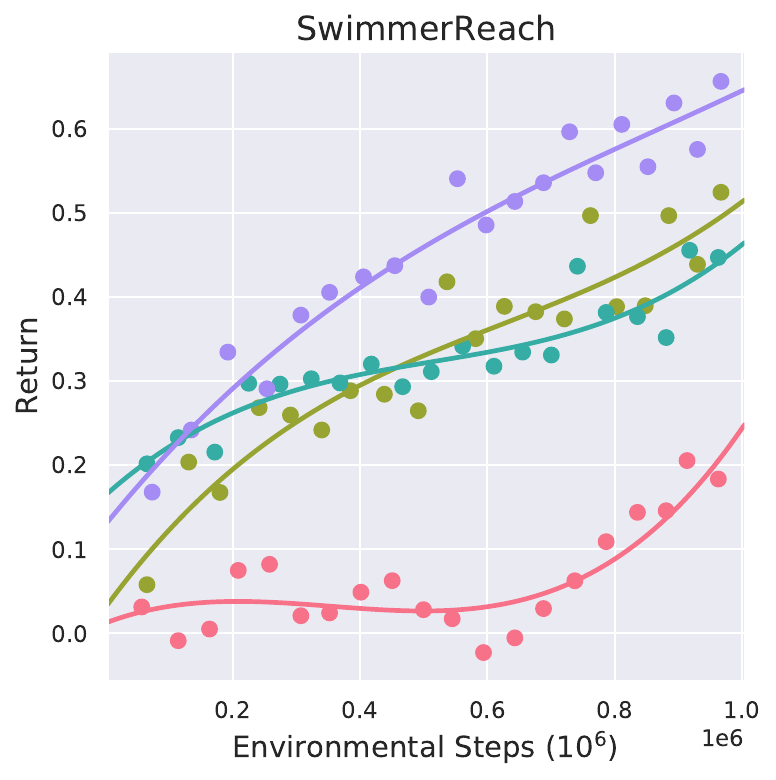}
  \includegraphics[height=3.4cm]{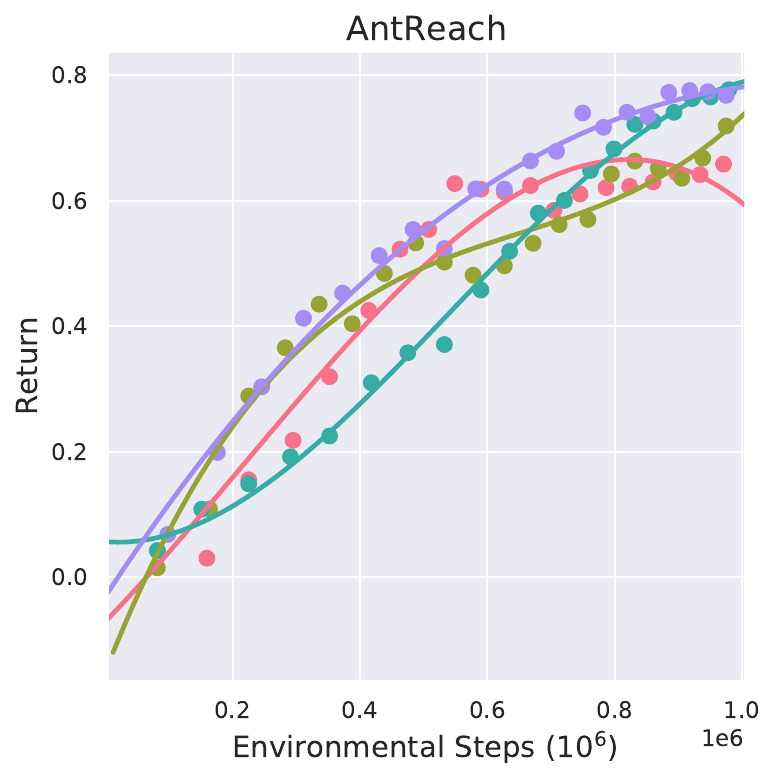}
  \includegraphics[height=3.4cm]{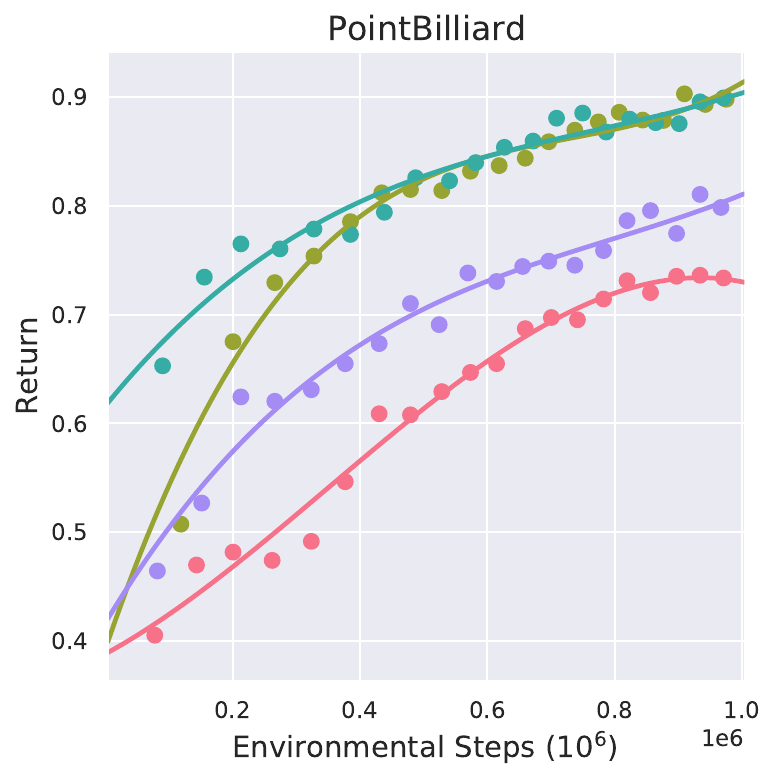}
  \includegraphics[height=3.4cm]{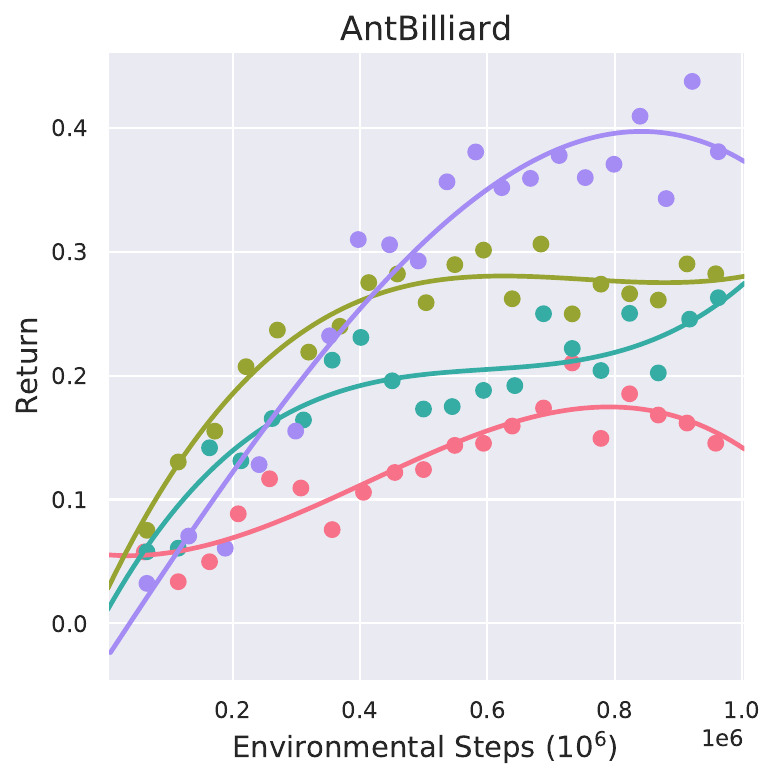}
  \includegraphics[height=3.4cm]{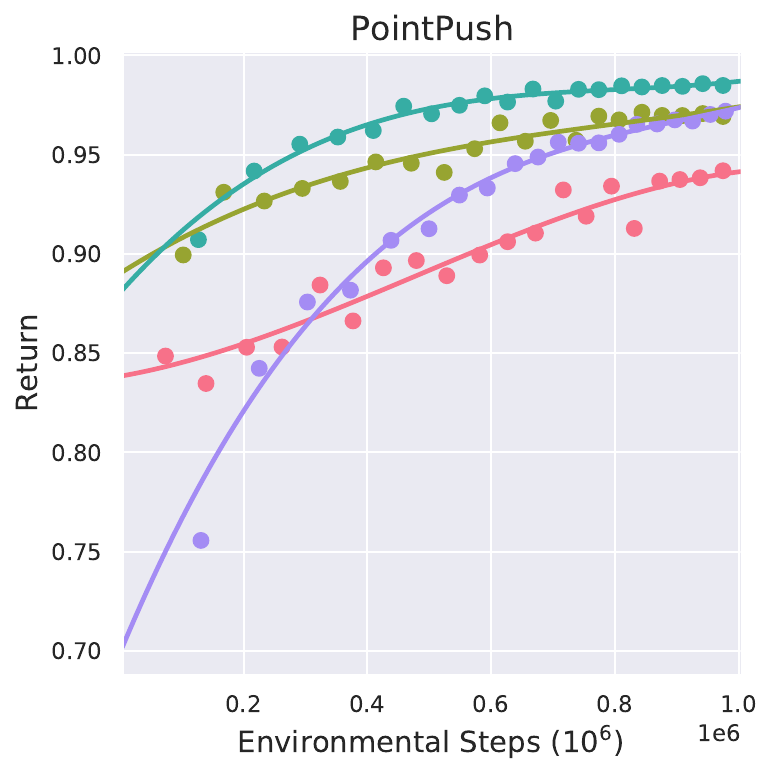}
  \includegraphics[height=3.4cm]{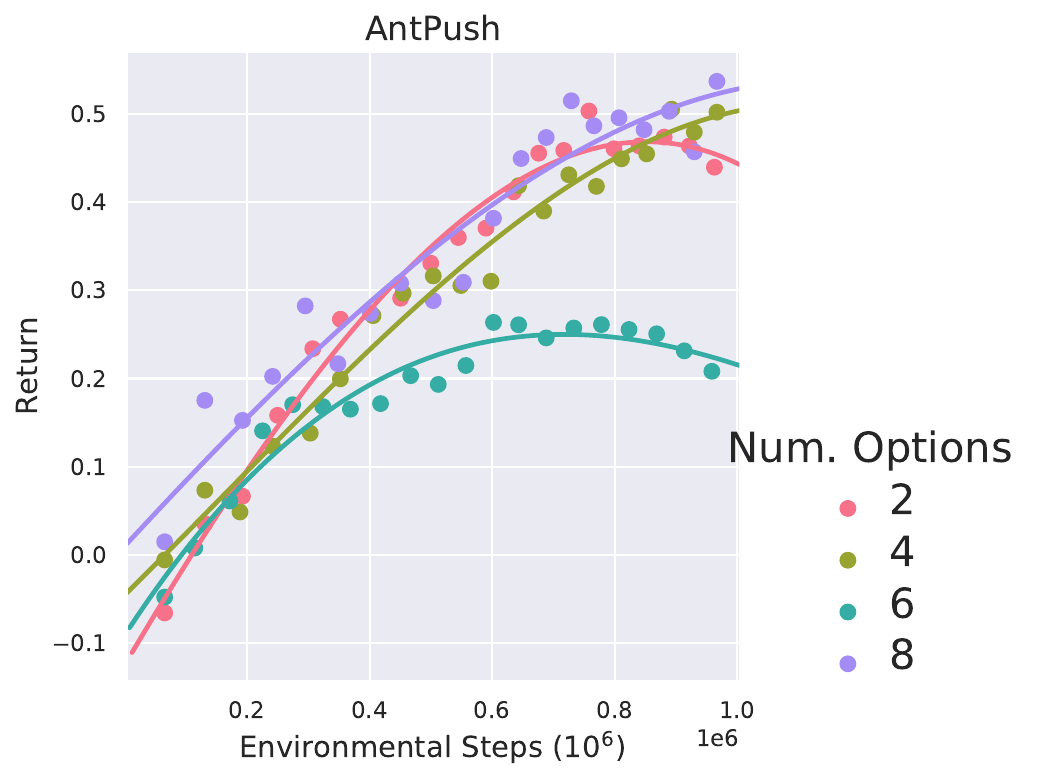}
  \vspace{-2mm}
  \caption{Learning curves for transfer learning experiments.}
  \label{figure:nopt-result}
  \vskip -0.1in
\end{figure}

\Cref{figure:nopt-result} shows learning curves of IMTC agents with variable number of options ($2, 4, 6, 8$) in task adaptation experiments.
Increasing the number of options sometimes improves the performance (e.g., in \texttt{SwimmerReach}), but the relationship is not obvious.
However, we can confirm that increasing the number of options does not have a bad effect on the performance, while it reduces training steps per each intra-option policy, thanks to network sharing \Cref{appendix:exp-details:nn}.
\end{document}